\documentclass{article}

\usepackage{arxiv}

\usepackage[utf8]{inputenc} 
\usepackage[T1]{fontenc}    
\usepackage{hyperref}       
\usepackage{url}            
\usepackage{booktabs}       
\usepackage{amsfonts}       
\usepackage{nicefrac}       
\usepackage{microtype}      
\usepackage{lipsum}
\usepackage{graphicx,makecell,amsmath}
\graphicspath{ {./images/} }

\title{Evolution of Meta's LLaMA Models and Parameter-Efficient Fine-Tuning of Large Language Models: A Survey}

\author{
 Abdulhady Abas Abdulla \\
  University of Kurdistan Hewler\\
   \And
 Arkaitz Zubiaga \\
  Queen Mary University\\
  \And
 Seyedali Mirjalili \\
  Torrens University Australia\\
   \AND
   Amir H. Gandomi \\
   University of Technology Sydney \\
   \And
   Fatemeh Daneshfar \\
   University of Kurdistan Sanandaj, Iran \\
   \texttt{daneshfarshadi@gmail.com} \\
   \And
   Mohammadsadra Amini \\
   TU Dortmund University\\
\And
   Alan Salam Mohammed \\
   University of Kurdistan Hewler \\
\And
   Hadi Veisi \\
   Tehran University \\
}

\begin{document}
\maketitle
\begin{abstract}
Abstract: This review surveys the rapid evolution of Meta AI's LLaMA (Large Language Model Meta AI) series - from LLaMA 1 through LLaMA 4 and the specialized parameter-efficient fine-tuning (PEFT) methods developed for these models. We first describe the LLaMA family of foundation models (7B-65B to 288B parameters), their architectures (including native multimodal and Mixture-of-Experts variants), and key performance characteristics. We then describe and discuss the concept of PEFT, which adapts large pre-trained models by updating only a small subset of parameters, and review five PEFT methods that have been applied to LLaMA: LoRA (Low-Rank Adaptation), LLaMA-Adapter V1 and V2, LLaMA-Excitor, and QLoRA (Quantized LoRA). We discuss each method's mechanism, parameter savings, and example application to LLaMA (e.g., instruction tuning, multimodal tasks). We provide structured discussion and analysis of model and adapter architectures, parameter counts, and benchmark results (including examples where fine-tuned LLaMA models outperform larger baselines). Finally, we examine real-world use cases where LLaMA-based models and PEFT have been successfully applied (e.g., legal and medical domains), and we discuss ongoing challenges and future research directions (such as scaling to even larger contexts and improving robustness). This survey paper provides a one-stop resource for ML researchers and practitioners interested in LLaMA models and efficient fine-tuning strategies. 
\end{abstract}


\section{Introduction}\label{sec1}
Large language models (LLMs) have become the central abstraction for modern natural language AI, unifying tasks such as understanding, long-form generation, dialogue, coding, and multi-modal reasoning under a single, pretrain-then-adapt paradigm \cite{1}\cite{2}. This shift has been driven primarily by the Transformer architecture, massive web-scale pre-training, and steady increases in computer and data that translate via scaling laws into predictable gains in sample efficiency and downstream performance. As models have grown, they have acquired strong zero/few-shot generalization and increasingly robust planning and tool-use behaviors. At the same time, extensions to longer context windows and multimodality have pushed capabilities beyond text-only I/O toward document reasoning, grounding, and vision-language tasks. At the same time, alignment and instruction-following fine-tunes have made these systems usable for interactive applications by steering pretrained knowledge into helpful, safe, and concise behavior for non-expert users \cite{1}. Within this landscape, Meta AI's LLaMA series has played an influential role in this landscape \cite{2}. The original LLaMA model (released February 2023) comprised open-source ``Foundation" Transformer models ranging from 7 billion (7B) to 65B parameters \cite{3}. Despite having fewer parameters than contemporaries like GPT-3 (175B) and PaLM (540B), LLaMA-13B matched or exceeded GPT-3 on many benchmarks \cite{2}. Following this, LLaMA 2 (July 2023) expanded the family to models up to 70B parameters, with specialized chat versions (LLaMA-2-Chat) fine-tuned for dialogue \cite{4}. Most recently, LLaMA 3 (2023-2024) introduced even larger and multimodal variants, including text-only models up to 405B (3.1 series) and vision-capable models (3.2 series) with 1B-90B parameters \cite{5}. In April 2025, Meta released the first members of LLaMA 4  Scout and Maverick, which use a sparse Mixture-of-Experts (MoE) architecture to achieve effectively trillions of parameters (17B active with many experts, distilled from a 288B flagship model) while supporting unprecedented 10-million token context windows. This explosive growth in size and context has greatly expanded the capabilities of LLaMA models \cite{6}.

However, fine-tuning such large models for downstream tasks by updating all weights (full fine-tuning) is often impractical due to computational and storage constraints. Instead, parameter-efficient fine-tuning (PEFT) methods have become essential \cite{7}. PEFT strategies freeze the majority of pre-trained models' parameters and introduce only a small number of adapter parameters (or modifications) that are trained on target tasks. This drastically reduces memory usage and training time while retaining performance close to full fine-tuning \cite{8}. Classic PEFT techniques include adapters, prefix-tuning, prompt tuning, and notably LoRA (Low-Rank Adaptation) \cite{9}. Recently, new PEFT methods have been designed specifically for LLaMA and related multi-modal instruction models, such as LLaMA-Adapter. \cite{10}, LLaMA-Adapter v2 \cite{11} and LLaMA-Excitor \cite{12}, as well as quantized training approaches like QLoRA \cite{13}.

\begin{figure}[!ht]
\centering
\includegraphics{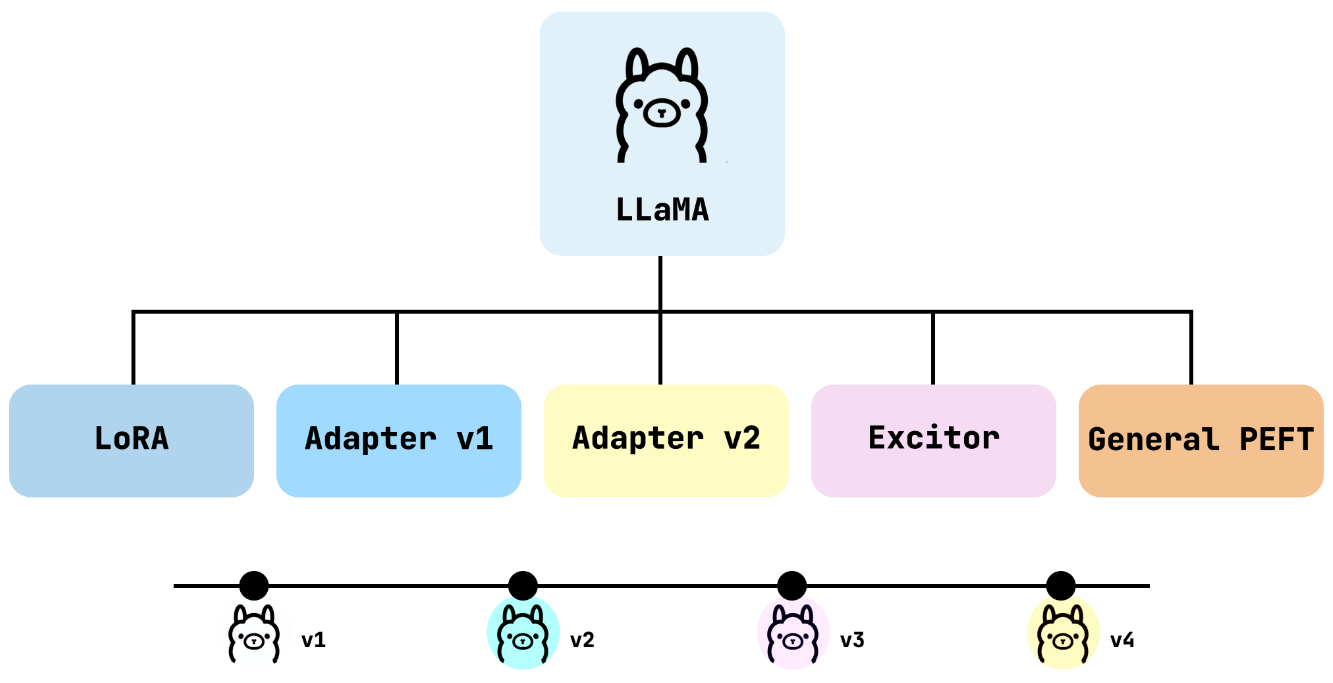}
\caption{Evolution of Meta's LLaMA models from foundation models (7B-65B) through chat-optimized variants to sparse Mixture-of-Experts (MoE) architectures}
\label{fig1}
\end{figure}

\begin{figure}[!ht]
\centering
\includegraphics{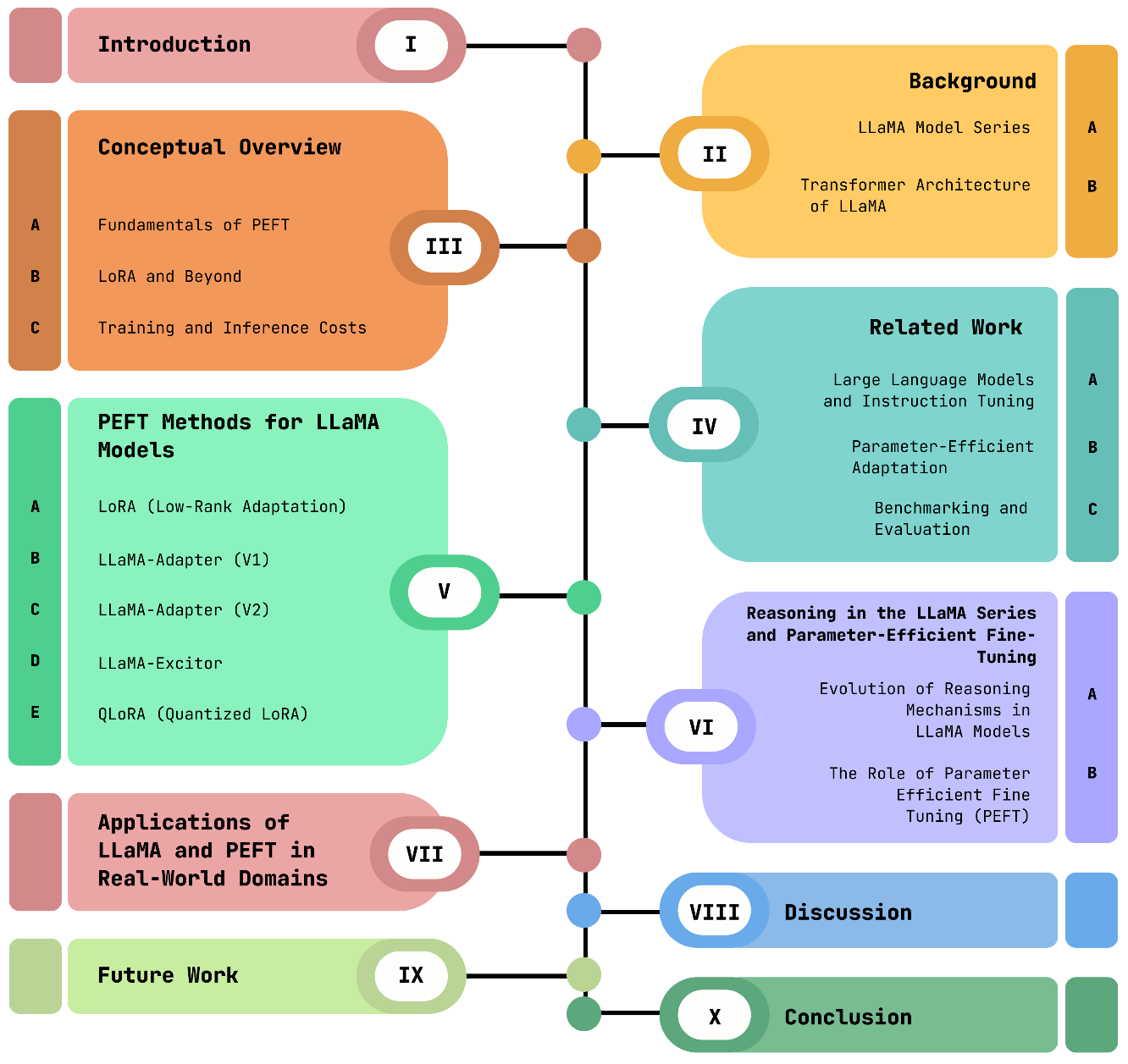}
\caption{Flowchart of the Survey Structure for LLaMA and Parameter-Efficient Fine-Tuning Methods}
\label{fig2}
\end{figure}

This paper reviews both the evolution of LLaMA models and the development of PEFT methods tailored to them, as summarized in Figure \ref{fig1}. We begin by summarizing the LLaMA model series and their architectural innovations (e.g., high-context, multi-modality, MoE) in the Background. We then provide a Conceptual Overview of these models and the general need for PEFT. In Related Work, we situate LLaMA within the broader context of LLM development and survey established fine-tuning approaches. In the PEFT Methods section, we examine each major method (LoRA, LLaMA-Adapter V1/V2, LLaMA-Excitor, QLoRA) in detail, explaining their mechanisms and how they apply to LLaMA. We also present Example Architectures, including diagrams and tables of model and adapter structures, parameter counts, and training regimes. We then discuss Applications where LLaMA and these PEFT techniques have yielded gains in real-world domains (e.g., medical and legal text processing, vision tasks) with quantitative results. Finally, we offer a Discussion of the state of the art and identify Future Work directions (such as further reducing fine-tuning overhead, improving training on noisy instruction data, and exploring new LLaMA extensions). We conclude by summarizing key insights. Throughout, we include tables and figures to illustrate model architectures, parameter efficiency comparisons, and benchmark outcomes. Existing LLM surveys synthesize the field broadly covering architectures, training, alignment, and applications across many model families (e.g., GPT/PaLM/LLaMA), but they do not center the analysis on a single open-weights lineage. By contrast, we narrow the lens to LLaMA: we trace the evolution of Meta's LLaMA family and tie it explicitly to LLaMA-specific PEFT recipes and multimodal adapters, with consolidated tables/figures on parameter scales, context lengths, MoE design, and empirical outcomes.  In scope and depth, this manuscript is, to our knowledge (as of October 2025), the first comprehensive LLaMA-centric survey that couples the model lineage (LLaMA 1 to 4) with a systematic, practitioner-oriented treatment of PEFT techniques tailored to LLaMA's architectures and deployment realities. In addition, the summary structure paper is shown in Figure \ref{fig2}.
 
\section{Background}\label{sec2}
\subsection{LLaMA Model Series}
\begin{figure}[!ht]
\centering
\includegraphics{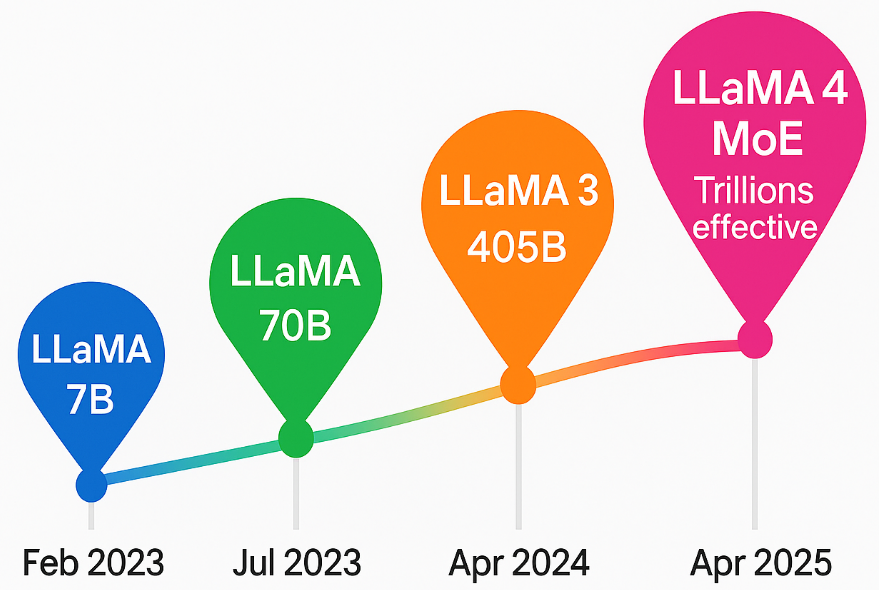}
\caption{LLaMA Model Scaling Timeline: From 7B to Trillions (2023-2025)}
\label{fig3}
\end{figure}

The LLaMA (Large Language Model Meta AI) family is a Transformer-based language model developed by Meta AI \cite{2}. The original LLaMA (also called LLaMA 1), introduced in early 2023, consisted of four text-only models with 7B, 13B, 33B, and 65B parameters \cite{3}, as shown in Figure \ref{fig3}. All were trained on publicly available text corpora and demonstrated that smaller parameter counts could achieve performance on par with much larger models. For instance, LLaMA-13B outperformed GPT-3 175B on most benchmarks, and LLaMA-65B was competitive with Chinchilla-70B and PaLM-540B \cite{5}. These models used the standard Transformer decoder architecture, trained with 2048-token context windows.
 
In mid-2023, Meta released LLaMA 2 \cite{4}. This version included new pre-trained models of 7B, 13B, and 70B parameters (text-only), as well as fine-tuned ``chat" variants (LLaMA-2-Chat) optimized for dialogue. LLaMA-2 models were trained on more data and incorporated safety mitigations. The 70B version significantly improved zero-shot capabilities and open-ended generation. The LLaMA-2 paper reports that these models ``outperform open-source chat models on most benchmarks" and represent an ``open-weights alternative" to closed-source LLMs. Later in 2023, LLaMA 3 \cite{5} introduced even larger and more diverse models. The LLaMA-3.1 series included text-only models of 8B, 70B, and 405B parameters (some with 128K token context). Alongside, the LLaMA-3.2 series introduced multimodal vision-capable models: 1B and 3B parameter text-only variants for low-latency inference, and 11B and 90B parameter models that accept both image and text inputs. All LLaMA 3 models support extremely long contexts (128,000 tokens). For example, the AWS Bedrock announcement notes that LLaMA 3.2's 90B ``Vision" model uses a 128K context and can perform advanced image-text reasoning, while the 11B Vision model is designed for efficient multimodal generationa. The smaller 3B and 1B models are tailored for edge devices and low-power fine-tuning. Additionally, the LLaMA-3.3 model (released Dec 2024) is a 70B parameter text model instruction-tuned for multilingual dialogue (supporting 8 languages, 128K context).

The most recent expansion is LLaMA 4 \cite{6}. In April 2025, Meta unveiled LLaMA 4 Scout and Maverick. Both have 17B active parameters organized in a sparse Mixture-of-Experts (MoE) framework, meaning they each route tokens through subsets of many ``experts" sub-networks. LLaMA 4 Scout has 16 experts, while LLaMA 4 Maverick has 128 experts. Despite only 17B active parameters, the models effectively incorporate the capacity of much larger networks (Maverick is distilled from a 288 B-parameter base). Critically, these models support a 10-million-token context window orders of magnitude beyond prior LLMs. Early analyses show LLaMA 4 Scout running inference on a single NVIDIA H100 GPU, and LLaMA 4 Maverick achieving state-of-the-art reasoning/coding performance among open models. The full LLaMA 4 Behemoth (still in training) is rumored to have 288B active parameters and $\sim 2$ trillion total (dense equivalent) parameters.

Table \ref{tbl1} provides an overview of LLaMA models (parameter counts, context sizes, modalities). All LLaMA 3 and 4 models greatly increase context length compared to LLaMA 1/2, and LLaMA 4 introduces MoE architecture. Each LLaMA version was released under progressively more permissive licenses. LLaMA 1 was available only for research, whereas LLaMA 2 introduced a "Community License" allowing broader use. LLaMA 4 is also available (non-commercial/community license) with model weights. The open availability of these large models has spurred a wave of research and application development, but also highlighted the need for efficient fine-tuning methods that respect computational constraints.
\newpage
\begin{table}[!ht]
\caption{Key characteristics of the LLaMA model series, including model sizes, context window lengths, supported modalities, and notable architectural features}
\label{tbl1}
\centering
\begin{tabular}{lcccp{4cm}}
\hline
\textbf{Version}	&\makecell{\textbf{Sizes}\\ \textbf{(Parameters)}}&	\makecell{\textbf{Context}\\ \textbf{Window}}&	\textbf{Modality}&	\textbf{Notes/Architecture}\\
\hline
\makecell[l]{LLaMA 1 \\(Feb 2023)}	&7B, 13B, 33B, 65B	&2K (approx.)&	Text only&	\makecell[l]{Standard decoder Transformer; \\foundation LLMs.}\\
\makecell[l]{LLaMA 2 \\(Jul 2023)}&	7B, 13B, 70B&	$\sim 2K	$&Text only / Chat	&Pretrained + instruction fine-tuned (Chat); improved data.\\
\makecell[l]{LLaMA 3.1\\ (2023)}&	8B, 70B, 405B&	128K&	Text only&	Larger language models; expanded training data.\\
\makecell[l]{LLaMA 3.2\\ (Nov 2023)}&
	\makecell{1B, 3B (text-only); \\11B, 90B (vision)}&	128K&	Text + Image (Vision)&	Multi-modal vision-language models; early fusion of image tokens.\\
\makecell[l]{LLaMA 3.3 \\(Dec 2024)}&	70B (instruct)	&128K&	Text only (dialogue)&	Instruction-tuned for dialogue (8 languages).\\
\makecell[l]{LLaMA 4 Scout \\(Apr 2025)}&
	\makecell{17B active\\ (16 experts)}&	
\makecell{10M \\(10 million)}&	Text + Image&	Mixture-of-Experts (MoE) sparse model; distilled from LLaMA-4 Behemoth.\\
\makecell[l]{LLaMA 4 Maverick\\ (Apr 2025)	}&
\makecell{17B active \\(128 experts)}&	10M	&Text + Image&	MoE model (many experts) for enhanced reasoning; distilled from 288B Behemoth.\\
\makecell[l]{LLaMA 4 Behemoth\\ (coming)}&
	\makecell{288B active \\($\sim 2T$ total)}&	$\sim 10M$&	Text + Image&	Flagship model (in training) with $\sim 320$ experts expected.\\
\hline
\end{tabular}
\end{table}

\subsection{Transformer Architecture of LLaMA}
\begin{figure}[!ht]
\centering
\includegraphics[scale=.5]{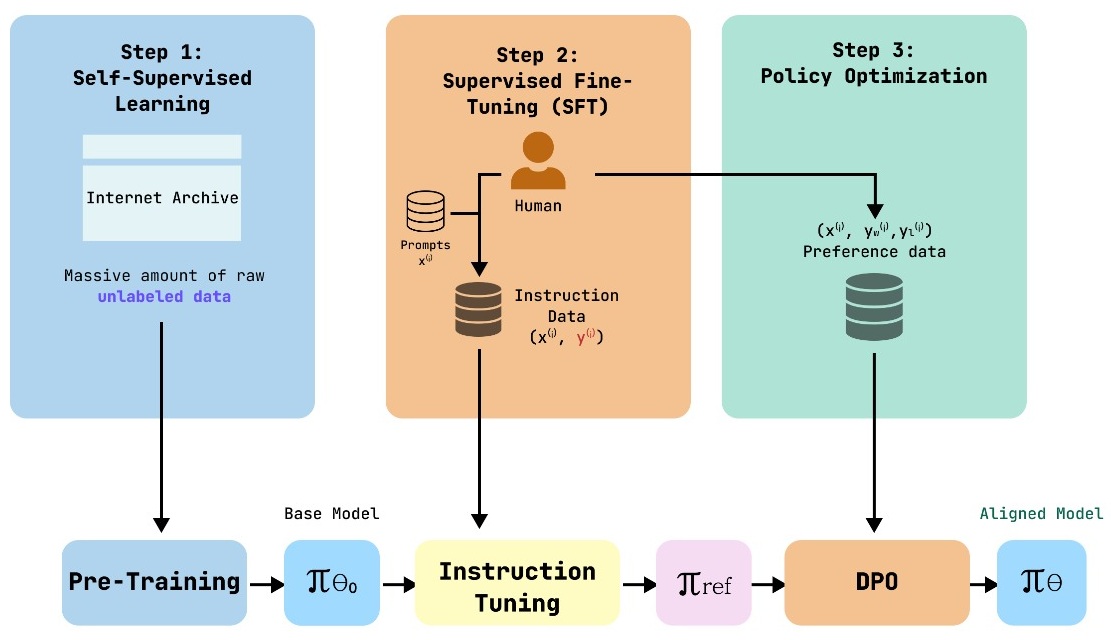}
\caption{Overview of the Training Pipeline Combining SFT and DPO}
\label{fig4}
\end{figure}

All LLaMA models are based on the standard Transformer decoder architecture. Each model consists of an embedding layer, multiple self-attention + feed-forward blocks, and a final projection to vocabulary logits \cite{2}. In detail, an input token sequence (possibly including special prompt tokens is first embedded and combined with positional encodings (RoPE in LLaMA1/2, extended context embeddings for LLaMA3). Each Transformer block applies multi-head self-attention and a two-layer MLP (feed-forward network) with a GELU or similar activation. The LLaMA papers do not radically alter the basic block but scale its width and depth for larger model sizes. The critical innovations in LLaMA 3/4 lie not in the Transformer block itself but in how it is scaled (e.g. more layers, larger hidden size, sparse experts) and how context is processed. Figure \ref{fig4} illustrates end-to-end pipeline for Supervised Fine Tuning (SFT) and Direct Preference Optimization (DPO) applied to these Transformer backbones \cite{14}. We begin by collecting prompts and generating multiple candidate outputs per prompt, then filtering them via rejection sampling guided by pairwise per-capability preference annotations. Those annotations serve both to train a reward model and to curate two SFT datasets, one general and one specialized. The SFT model, which retains the original Transformer and MoE blocks frozen, is trained on this data, and its best variants are carried forward alongside top performers from previous rounds. Finally, DPO refines these candidates using the specialized SFT data to produce the final optimized model. By layering this structured preference-driven workflow atop the standard Transformer decoder, including sparse experts in LLaMA 4, we achieve substantial gains without altering the core architecture.
 
Notably, LLaMA 4 introduces sparse Mixture-of-Experts (MoE) layers. In a MoE layer, instead of a single feed-forward network (FFN), there are many ``expert" FFNs. A learned router (or gating network) chooses which expert(s) each token attends to. Figure \ref{fig1} (below) illustrates a Switch Transformer-style MoE block, like those used in LLaMA 4. Two input tokens, ``More" and ``Parameters" are routed to different FFN experts by the router, after which their outputs are recombined. This allows the model to effectively have a much larger capacity (e.g. 128 experts $\times$  a given hidden size) while only performing computation in a few experts per token \cite{6}.
  
\begin{figure}[!ht]
\centering
\includegraphics[scale=.9]{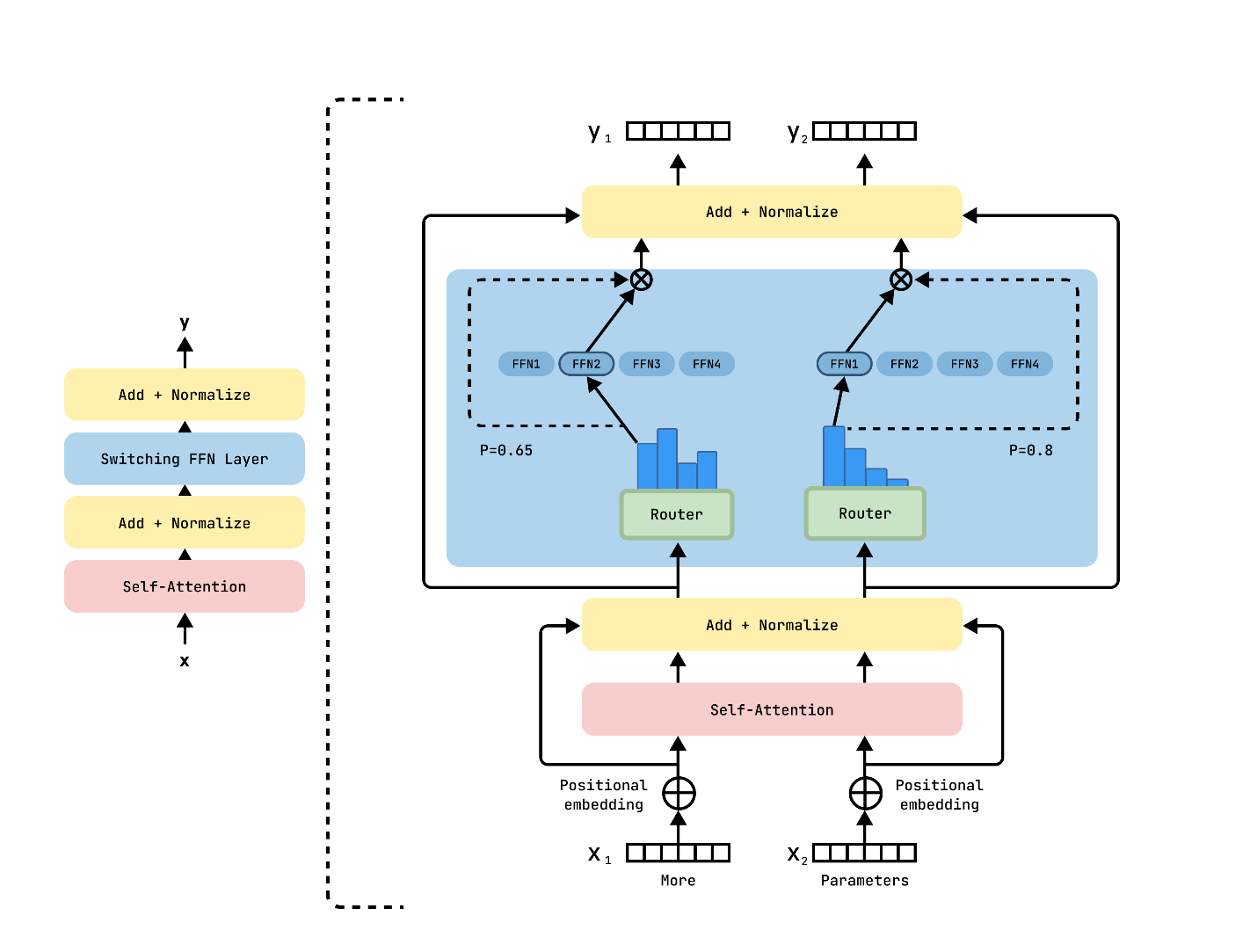}
\caption{Sparse Mixture-of-Experts Transformer block}
\label{fig5}
\end{figure}

In Figure \ref{fig5}, the Mixture-of-Experts (MoE) Transformer layer (based on the Switch Transformer concept) is shown. In a sparse MoE block, a routing network directs each input token $(x_1, x_2)$ to a subset of experts. Here token ``More" $(x_1)$ is sent to Expert 2 with weight 0.65, and ``Parameters" $(x_2)$ is sent to Expert 1 with weight 0.80. Only the selected expert FFNs ($FFN_2$ for $x_1$, $FFN_1$ for $x_2$) are applied. This expands model capacity without increasing per-token computation by having many experts in total but activating only a few per token. The introduction of MoE makes LLaMA 4 capable of training far larger models (effectively trillions of parameters) within practical computational budgets. However, it also poses new challenges for fine-tuning, as the routing network and expert parameters can complicate adaptation. We will revisit MoE when discussing how PEFT methods work (see Section \ref{sec5}).
\section{Conceptual Overview}\label{sec3}
This section motivates why full fine-tuning of large LLaMA models is impractical and introduces parameter-efficient alternatives. It sketches the core mechanisms, adapters, prompt/prefix tuning, LoRA/QLoRA, showing how small add-ons modify Transformer blocks with minimal trainable weights, and highlights the resulting cost and deployment trade-offs to orient the methods that follow.
\subsection{Fundamentals of PEFT}
Fine-tuning large pre-trained models is traditionally done by updating all parameters on downstream data, but this becomes infeasible as model sizes grow.  Parameter-Efficient Fine-Tuning (PEFT) tackles this by freezing the pre-trained model's original weights and introducing only a small set of task-specific parameters \cite{8}. Formally, if the base model has N parameters, PEFT adds just $M\ll N$ learnable parameters, often well under 1\% of N-yet achieves performance on par with full fine-tuning. By doing so, it dramatically reduces both GPU memory footprint and computational cost during training and makes it practical to store multiple lightweight adapters for different tasks \cite{15}. PEFT methods work by injecting small modules or masks into the network. For example, adapters can be extra small dense layers inside Transformer blocks; prompt-tuning adds a few ``prompt" tokens to the input; prefix-tuning prepends learnable vectors at each layer's input; LoRA injects low-rank update matrices into key weight matrices. These approaches share the philosophy of leaving most of the model unchanged. Importantly, PEFT can often be merged into the base model after training (especially LoRA), so there is no inference overhead \cite{9}. As \cite{15} note in a recent survey, PEFT ``provides a practical solution" to LLM adaptation by ``minimizing the number of trainable parameters". When LLaMA models (some with hundreds of billions of parameters) are deployed, PEFT is not just convenient but often necessary to fine-tune them on new data. Indeed, many research groups rapidly adopted LoRA and related techniques to fine-tune LLaMA for specialized tasks (chatbot behavior, code generation, etc.).
\subsection{LoRA and Beyond}
LoRA (Low-Rank Adaptation) works by keeping the original pre-trained weight matrix $W_0\in R^{d\times k}$ fixed and learning only a small, low-rank correction \cite{16}. Specifically, it expresses the update as
\begin{equation}
 \Delta W=\frac{\alpha}{r} BA 
\end{equation}
where $B\in R^{d\times r}$ and $A\in R^{r\times k}$ are the only trainable parameters (with $r\ll\min(d,k)$ ), and $\alpha$ is a scaling factor (often chosen so that $\alpha⁄r=1$). The adapted weight is then
\begin{equation} 
W=W_0+\Delta W 
\end{equation}
Because only $B$ and $A$ (a total of $dr+rk$ parameters) are updated, LoRA can reduce the number of trainable parameters by several orders of magnitude while retaining full model capacity at inference time.

Beyond LoRA, more specialized adapter techniques have been proposed that better suit instruction-tuning or vision-language tuning. For LLaMA specifically, LLaMA-Adapter \cite{10} and LLaMA-Excitor \cite{12} are novel PEFT methods designed to adapt LLaMA into an instruction-following model or visual instruction model, using a combination of prompt embeddings and modified attention. We will detail these in Section \ref{sec5}. Another recent innovation is QLoRA (Quantized LoRA), which combines LoRA with 4-bit quantization to enable fine-tuning of 65B models on a single 48GB GPU.
\subsection{Training and Inference Costs}
The practical benefits of PEFT are clear in terms of resource savings. Full fine-tuning a model like LLaMA-70B (70 billion parameters) would require loading all weights in GPU memory and computing gradients for them, which demands tens to hundreds of GB of GPU RAM and long training times. In contrast, LoRA or adapter-based fine-tuning can often be done on a few GPUs or even a single high-memory GPU. For example, \cite{10} reports that LLaMA-Adapter could fine-tune a 7B model on 8 A100 GPUs in under one hour, whereas Alpaca (fully fine-tuned LLaMA-7B) required several GPU-days. QLoRA goes even further: it uses a frozen 4-bit quantized LLaMA with only LoRA adapters, allowing 65B models to be fine-tuned on a single 48GB GPU in $\sim 24$ hours \cite{16}. Crucially, these PEFT adapters can be stored separately (usually only a few MB for small ranks) and merged at inference time. In deployment, the added latency is minimal or zero (if merged). In summary, PEFT makes it tractable to use giant LLaMA models in practice by cutting fine-tuning cost by orders of magnitude. We now survey related LLM work for context, before diving into the specifics of LLaMA-oriented PEFT techniques.
\section{Related Work}\label{sec4}
\subsection{Large Language Models and Instruction Tuning}
LLaMA models join a growing family of large, Transformer-based language models initiated by GPT-3. Over the past few years, models like PaLM (540B) \cite{17}, Chinchilla, OPT, and others have established that scale and data are key to performance \cite{18}. Many recent models incorporate instruction tuning, where a base model is fine-tuned on human or synthetic instruction-answer pairs to follow commands (e.g. FLAN, Alpaca). Notable examples include Stanford's Alpaca (fine-tuned LLaMA-7B on 52K synthetic instructions) and CarperAI's Vicuna (fine-tuned LLaMA-13B on chat data). The open-release trend of LLaMA spurred many of these; in particular, Alpaca used LLaMA as the base model to democratize instruction-following capabilities. Beyond NLP, multi-modal and vision-language models like CLIP \cite{19}, DALL·E \cite{20} , and BLIP \cite{21} have influenced LLaMA's multimodal branches. Some works directly combine vision backbones with LLaMA; for instance, LLaVA adds a visual transformer encoder to LLaMA for answering visual questions \cite{22}. The LLaMA-3.2 Vision models align images with text inputs natively, reducing the need for separate encoders.
\subsection{Parameter-Efficient Adaptation}
The idea of keeping most parameters fixed, and tuning only a small subset, was explored early in NLP. Adapter modules \cite{23} inserted small bottleneck layers into Transformers. Prefix-tuning and prompt-tuning \cite{24}prepended trainable vectors to the input or each layer. LoRA \cite{9} introduced trainable low-rank matrices on attention weights. Each of these methods was shown to require far fewer trainable parameters than full fine-tuning, with often negligible loss of performance. More recent surveys show that combining PEFT with quantization or pruning can further reduce resource needs. Contemporary to LLaMA, many models have been fine-tuned with PEFT. For example, numerous fine-tuning recipes for GPT-3 \cite{18} and PaLM use LoRA to adapt them to specific tasks with limited compute. The Hugging Face community has embraced PEFT widely, providing libraries for LoRA, QLoRA, and adapters.
\subsection{Benchmarking and Evaluation}
Performance of fine-tuned LLMs is measured on tasks like question answering, summarization, code generation, and instruction-following \cite{18}. Benchmarks such as MMLU \cite{25} (multiple-choice knowledge tests), AlpacaEval \cite{26} (instruction response quality), and GPT-4's evaluations are used. For multi-modal models, tasks include COCO Captioning \cite{27} and ScienceQA \cite{28}. As we will see, both LLaMA models and PEFT variants are evaluated on these benchmarks. For instance, LLaMA-Adapter's \cite{10} vision variant achieved a state-of-art CIDEr score on MSCOCO. LLaMA-Excitor notably improved MMLU by 6\% compared to baselines.

Overall, prior work has established that carefully chosen PEFT methods allow large models to be adapted for new tasks with minimal overhead. Our focus is to bring together this paradigm specifically for the LLaMA family and its ecosystem.
\section{PEFT Methods for LLaMA Models}\label{sec5}
In this section, we examine five key PEFT methods that have been applied to LLaMA models. For each, we describe the approach, its advantages, and examples of applying it to LLaMA. Transformer Block (Base LLaMA): Below is a self-contained description of the Transformer decoder block as used in LLaMA (Figure \ref{fig3}a), with all key computations and connections spelled out.

 \begin{figure}[!ht]
\centering
\includegraphics{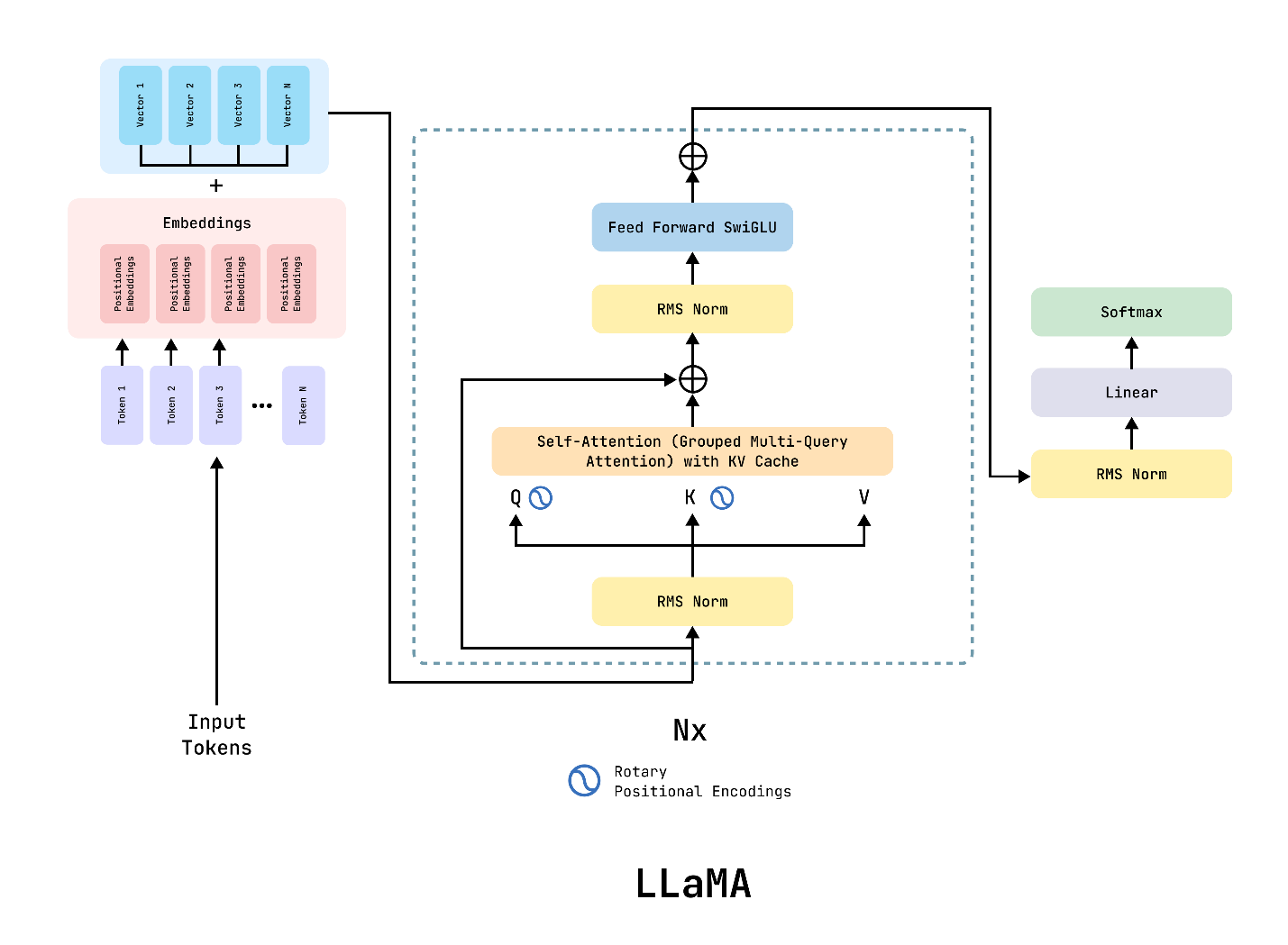}
\caption{Architecture of LLaMA Transformer}
\label{fig6}
\end{figure}

\begin{figure}[!ht]
\centering
\includegraphics[scale=.95]{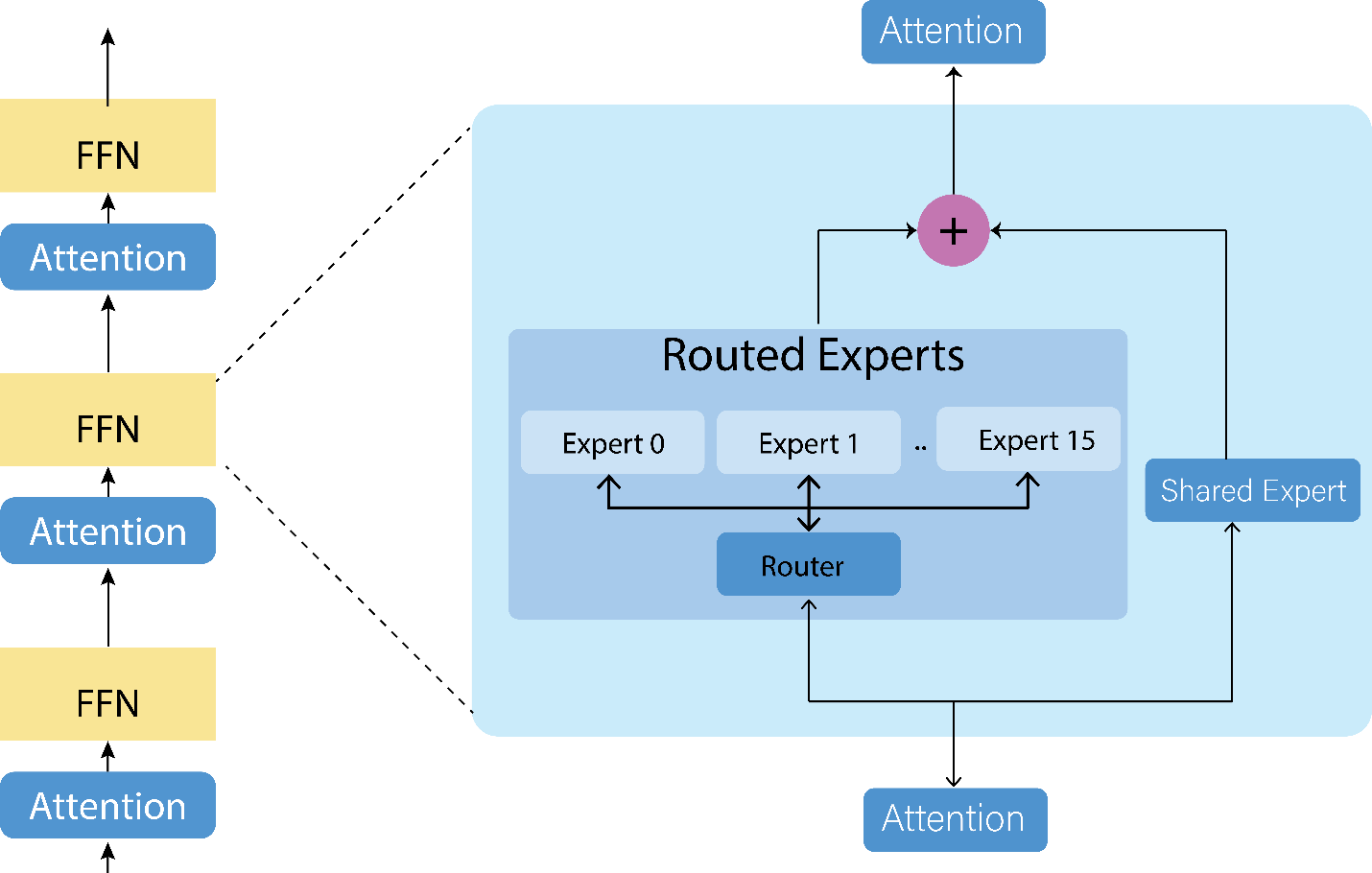}
\caption{Mixture-of-Experts Transformer Layer with Dynamic Expert Routing.}
\label{fig7}
\end{figure}

In LLaMA's decoder stack, as shown in Figure \ref{fig6}. Each block processes its input 
$X\in R^{L\times d}$ (where L is the sequence length and d the hidden size) through two sub-layers, multi-head self-attention and a position-wise feed-forward network (FFN), each wrapped with a residual ("Add") connection and layer normalization ("Norm"). 
\begin{enumerate}
\item	Multi-Head Self-Attention\\
First, three linear projections map X to queries, keys, and values:
\begin{equation} 
Q=W_Q X,K=W_K X,V=W_V X 
\end{equation}
where $W_Q,W_K,W_V\in R^{d_k\times d}$ and $d_k=d/num\_heads$. Next, scaled dot-product attention computes
\begin{equation} SA(X)=softmax\Bigl(\frac{QK^T}{\sqrt{d_k}} \Bigr)V\in R^{L\times d} 
\end{equation}
The block then adds this attention output back to the input and normalizes:
\begin{equation} 
Y=LayerNorm(X+SA(X)) 
\end{equation}
\item	Feed-Forward Network (FFN)\\
The normalized output $Y$ is passed through a two-layer MLP with a nonlinearity-typically GELU:
\begin{equation} 
FFN(Y)=W_2 (GELU(W_1 Y)) 
\end{equation}
where 
$W_1\in R^{d_{ff}\times d}$,
$W_2\in R^{d\times d_{ff}}$, 
and 
$d_{ff}$ (often 4d )
 is the inner-layer size. Again, a residual connection and layer norm follow:
\begin{equation} 
Z=LayerNorm(Y+FFN(Y)) 
\end{equation}
Putting it all together, one decoder block implements
\begin{align}
\begin{aligned}
&SA(X)&&softmax\Bigl(
\frac{W_Q X(W_{\kappa} X)^T}{\sqrt{d_k}}
\Bigr) W_V X\\
&Y&&LayerNorm(X+SA(X))\\
&Z&&LayerNorm(Y+W_2 (GELU(W_1 Y))) )
\end{aligned}
\end{align}
\end{enumerate}
In practice, each of these projections is split across multiple heads, and the resulting head outputs are concatenated back to dimension d. A full LLaMA model stacks $N$ such blocks (e.g. $\setminus N=32$ for a 7 B parameter model up to $N=96$ for larger ones), allowing deep contextualization while preserving stable gradients via the "Add \& Norm" pattern. The Llama 4 series builds on earlier releases by delivering state-of-the-art multimodal capabilities at a fraction of the cost, outperforming many substantially larger models. Achieving this required several novel strategies during pre-training \cite{6}. First, Llama 4 adopts a Mixture-of-Experts (MoE) architecture (see Figure \ref{fig7}). In this design, each token activates only a small subset of the model's total parameters, yielding significant compute savings without sacrificing quality. For example, the Maverick variant stores 400 B parameters but activates just 17 B at inference time. Dense and MoE layers are interleaved so that every token is processed by a shared expert plus one of 128 routed experts \cite{6}. Although all weights remain in memory, only that subset fires during a forward pass, which dramatically reduces serving cost and latency. As a result, Llama 4 Maverick can run on a single NVIDIA H100 DGX node or be distributed across multiple GPUs for even greater throughput. Llama 4 also integrates text and vision inputs early in the network, enabling joint pre-training on unlabeled text, images, and video. The vision encoder, based on MetaCLIP and trained alongside a frozen Llama backbone, was enhanced to better align visual representations with the language model. To stabilize training across diverse configurations, the MetaP procedure was developed to select per-layer learning rates and initialization scales reliably. These hyperparameters generalize across different batch sizes, model depths, and token budgets. The pre-training corpus exceeded 30 trillion tokens, which is ten times more multilingual data than Llama 3, covering 200 languages and over 100 languages with more than 1 billion tokens each. Finally, performance was optimized using FP8 precision, achieving 390 TFLOPs per GPU on 32 K GPU runs for the Behemoth variant. A subsequent mid-training phase with specialized datasets extended the context window and unlocked a best-in-class 10 million-token input length in Llama 4 Scout \cite{6}.
 
\subsection{LoRA (Low-Rank Adaptation)}\label{sec5.1}
LoRA (Hu et al., 2021) is a straightforward, yet highly effective PEFT method that freezes the pre-trained model weights and injects small, trainable low-rank adapters into selected layers. In Transformer blocks, LoRA modules are typically added to the query and value projection matrices (and optionally the MLP).  Below is a detailed description of how LoRA is injected into a single LLaMA decoder block (Figure \ref{fig3}b), preserving all the mathematical notation and parameter-efficiency details:
\paragraph*{\rm 1.	Original Query Projection}\hfill\\
The standard self-attention query projection is
\begin{equation} 
W_Q^{orig} \in R^{d_k\times d}
 \end{equation}
which maps the hidden states
 $X\in R^{L\times d}$ to queries 
$Q=W_Q^{orig} X$.
\paragraph*{\rm 2. Low-Rank Decomposition}\hfill\\
LoRA replaces 
$W_Q^{orig}$
with a sum of the frozen original plus a low-rank update:
\begin{equation} 
W_Q=W_Q^{orig} +BA,\quad A\in R^{r\times d},\quad B\in R^{d_h\times r} 
\end{equation}
\begin{itemize}
\item[a.] A and B are the only trainable parameters.
\item[b.]	The product BA has rank at most r, so it injects only an r-dimensional "delta" into the projection.
\end{itemize}
\paragraph*{\rm 3.	Shapes and Parameter Counts}\hfill\\
\begin{itemize}
\item[a.]	Hidden size $d_h\approx 4,096$ (in LLaMA-7B), input dimension $d=d_h$.
\item[b.]	With $r=8$, the total extra parameters are
\begin{equation} 
A\vee +B\vee r\cdot d+d_h\cdot r=8\times 4096+4096\times 8=65,536+32,768=98,304 \end{equation}
\end{itemize}
which is roughly 0.01\%  of the full model's $\sim 7$ billion parameters.
\paragraph*{\rm 4. Generalization to Other Projections}\hfill\\
Exactly the same decomposition 
$W=W^{orig} +BA$; can be applied to:
\begin{itemize}
\item[a.]
	Key and value projections $W_K,W_V$.
\item[b.]	The two linear layers in the FFN, $W_1$ and $W_2$.
\end{itemize}
\paragraph*{\rm 5.	Resulting Block}\hfill\\
Apart from replacing $W_Q^{orig}$  with $W_Q^{orig} +BA$ (and similarly for any other chosen weight matrices), The rest of the Transformer block multi-head attention mechanics, residual adds, LayerNorm, and FFN structure remain unchanged. This lets LoRA achieve effective adaptation with only a tiny fraction of additional parameters.

In the original paper, LoRA achieved up to a $10,000\times$  reduction in fine-tunable parameters on GPT-3 175B (using 4) while matching full-fine-tuning performance. This drastic memory saving enables fine-tuning very large models-e.g., adapting a 65B-parameter LLaMA on a single high-memory GPU, something infeasible with conventional backpropagation. LoRA has seen widespread adoption in LLaMA-based pipelines. For example, Clinical LLaMA-LoRA (Gema et al., 2023) applied a two-stage LoRA adaptation to LLaMA-7B on medical text and reported a 13.01\%  boost in macro-averaged AUROC over the unfine-tuned baseline, demonstrating LoRA's ability to specialize large models efficiently. When applying LoRA to LLaMA, practitioners must decide which layers to adapt and what rank r to choose. A common recipe is to insert LoRA into the self-attention's key/query/value projections (and sometimes the MLP), with r in the range 4-16. Smaller models often tolerate larger $r$ (e.g. $r=8$ or 16 for LLaMA-7B), whereas very large models may use smaller r to keep adapter size minimal. Often, LoRA-adapted models are then further fine-tuned on downstream tasks (e.g. instruction data) to maximize performance.
\subsection{LLaMA-Adapter (V1)}\label{sec5.2}
Large pre-trained language models (LLMs) such as LLaMA have demonstrated extraordinary capabilities across a wide range of tasks \cite{2}. However, fully fine-tuning these models for each new downstream task is prohibitively expensive in terms of GPU memory, compute time, and storage of multiple model variants. PEFT addresses this by freezing the bulk of the model's weights and only learning a small, task-specific subset of parameters. Early PEFT approaches LoRA, prefix-tuning, adapters show that updating just 0.1-1\% of parameters can match the performance of full fine-tuning, but stability and speed remain challenges, especially as model size scales.
 
\begin{figure}[!ht]
\centering
\includegraphics{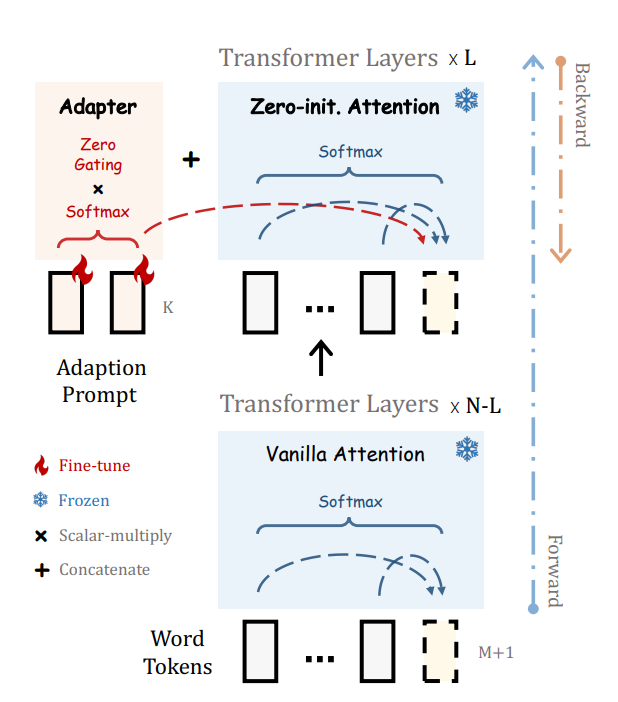}
\caption{Illustration of LLaMA-Adapter V1 \cite{10}.}
\label{fig8}
\end{figure}

LLaMA-Adapter \cite{10} builds on this paradigm with two core innovations:
\begin{enumerate}
\item
	Prompt-based adaptation inspired by prefix-tuning, which leverages learned ``soft prompts" inserted at every layer to steer behavior.
\item	Zero-init attention gating, which eliminates sudden perturbations to the pre-trained model by starting the adapter's influence at zero and learning a scalar gate that smoothly ``turns on" the new signals.
\end{enumerate}
This combination yields a highly stable, fast-converging method that injects new capabilities without disturbing the base model's original knowledge.

Layer-wise Prompt Insertion: At each Transformer layer l, we introduce a small set of m learnable prompt vectors:
\begin{equation}
 P^l=[p_1^l,p_2^l,\ldots,p_m^l ]\in R^{m\times d}, \end{equation}
where d is the hidden dimension. These prompts are applied to the incoming hidden states 
$H^{l-1}\in  R^{n\times d} $:
\begin{equation} 
X^l=\begin{bmatrix}
P^l\\
H^{l-1} 
\end{bmatrix}
\in R^{(m+n)\times d}.
 \end{equation}
Because prompts are learned parameters, they can encode rich signals to guide the layer's computations toward the new task.

 Zero-Initialized Attention Gating: The adapter computes an attention update $\Delta^l$ over the extended sequence $X^l $:
\begin{equation} 
\Delta ^l=softmax\Bigl(
\frac{X^l W_Q (X^l W_K )^{\top}}{\sqrt{d}}
\Bigr) X^l W_V\in R^{(m+n)\times d}. 
\end{equation}
To prevent any abrupt change to the pre-trained model, a layer-specific scalar gate $\lambda^l$ (initialized to zero) modulates this update:
\begin{equation} 
H^l=H^{l-1}+\lambda^l \Delta^l, 
\lambda^l|_{\text{init}} =0 
\end{equation}
During training, $\lambda^l$ is learned alongside the prompts. At the outset, $\lambda^l=0$ ensures $\Delta^l$ has no effect, preserving the model's original behavior. As gradients flow, $\lambda^l$ gradually increases, "injecting" the adapter's influence in a controlled manner.

Parameter Accounting: Across all $L$ Transformer layers, the total number of additional parameters is
\begin{equation} 
\underbrace{L\times (m\times d)}_{\text{prompt embeddings}}+\underbrace{L}_{\text{gating scalars}} \approx 1.2 \text{million} 
\end{equation}
for a typical choice of $m=10$ prompts per layer and $d=4096$. This represents just 0.017\%  of LLaMA- 7 B's 7 billion parameters.
\subsection{LLaMA-Adapter V2}
\begin{figure}[!ht]
\centering
\includegraphics{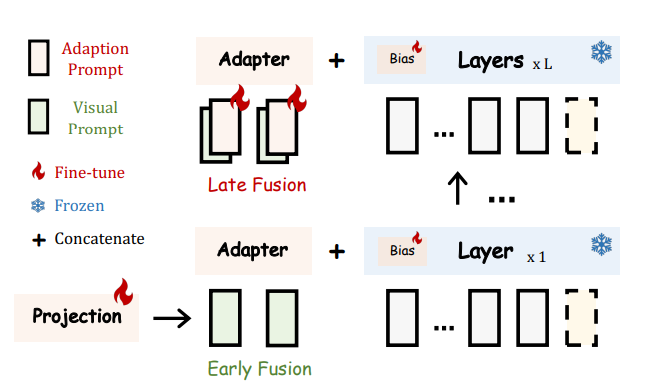}
\caption{Illustration of LLaMA-Adapter V2 \cite{11}.}
\label{fig9}
\end{figure}

LLaMA-Adapter V2 \cite{11} extends the V1 adapter to improve multi-modal and open-ended instructions, as shown in Figure \ref{fig9}.

While LLaMA-Adapter V1 focused on language-only prompts, Adapter V2 allows vision tokens and unlocks more learnable parameters. The main enhancements are:
\begin{itemize}
\item[a.]
	More trainable parameters: V2 ``unlocks" additional parameters in the model (e.g. layernorm scales, bias terms) beyond just the prefix prompts. This distributes the adaptation throughout the LLaMA model, not only in the adapter prompts. As a result, V2 introduces about 14 million parameters total (versus 1.2M in V1).
\item[b.]	Early fusion of vision tokens: Instead of feeding image features at the top layer, LLaMA-Adapter V2 inserts image token embeddings into \textit{early} Transformer layers. This allows the model to incorporate visual information more deeply.
\item[c.]	Joint training for vision and language: The model is trained on both image-text instructions and text-only instructions, with separate parameter groups. This avoids interference between modalities.
\item[d.]	Modular experts at inference: At inference time, Adapter V2 can optionally use off-the-shelf captioning or OCR experts to preprocess images, enhancing its visual understanding without retraining.
\end{itemize}
The result is a model that can handle open-ended multimodal instructions. Gao et al. report that LLaMA-Adapter V2 (with 14M parameters) surpasses the original Adapter and even GPT-4 on certain vision QA tasks. They also note that V2 strengthens language-only instruction-following beyond V1. In summary, Adapter V2 trades more adapter parameters (14M) for greater flexibility and multimodal reasoning. It remains vastly smaller than full finetuning a comparable multi-modal LLM. The V2 approach is particularly useful for extending LLaMA's reasoning abilities to image-based tasks.

Notation Let the base LLaMA model be denoted by
\begin{equation} 
f(x;\theta), 
\end{equation}
where $\theta$ are the frozen pretrained weights.
\begin{itemize}
\item
	Denote the adapter (trainable) parameter set by $\Delta \theta$, and the prefix is prompted by a matrix $P\in R^{L\times d}$.
\item	Total trainable parameters in V1:
\begin{equation} 
|\Delta \theta_{V1} |=P\vee \approx 1.2\times 10^6. 
\end{equation}
\item	Total trainable parameters in V2:
\begin{equation} 
|\Delta \theta_{V2} |=P\vee +|\Delta \theta_{\text{unlocked}}  |\approx 14\times 10^6. \end{equation}
\end{itemize}
\paragraph*{\rm 1.	More trainable parameters}\hfill\\
V2 "unlocks" additional components (layer-norm scales, biases, etc.), so that
\begin{equation} 
\Delta \theta_{V2}=\{P007D\}\cup \{\Delta \gamma^{(i)},\Delta b^{(i)} \}_{i-1}^N, \end{equation}
where N is the number of Transformer layers, and $\Delta \gamma^{(i)},\Delta b^{(i)}$ are the extra scale/bias offsets learned for layer i.
\paragraph*{\rm 2. Early fusion of vision tokens}\hfill\\
Represent an image v by a sequence of patch-embeddings
\begin{equation} 
E_{lmg}=\phi(v)\in R^{T\times d}. 
\end{equation}
Instead of concatenating $E_{img}$ only at the top, $V2$ injects it at an early layer $L_0$ :
\begin{equation} 
H^{(l)}=
\begin{cases}
\text{TransformerLayer}[H^{(l-1)} ],&l<L_0,\\
\text{TransformerLayer}[[H^{(l-1)};E_{lmg} ]],&l=L_0,\\
\text{TransformerLayer}[H^{(l-1)} ],&l>L_0.
\end{cases}
\end{equation}
\paragraph*{\rm 3.	Joint training for vision and language}\hfill\\
Define two datasets:
\begin{equation} 
D_{\text{text}}=\{(x_t,y_t )\},D_{\text{vision}} =\{(x_v,y_v )\}, 
\end{equation}
where $x_t$ is a text prompt, $x_v=(v, prompt)$ an image-text instruction. The adapter is learned by minimizing the combined loss:
\begin{equation} 
L(\Delta \theta)=
\sum_{(x_t,y_t )\subset D_{\text{test}}}
 l(f(x_t;\theta,\Delta \theta),y_t )+
\sum_{(x_v,y_v )\subset D_{\text{rubse}}} l(f(x_v;\theta,\Delta \theta),y_v ). 
\end{equation}
\paragraph*{\rm 4.	Modular experts at inference}\hfill\\
At test time, one can optionally preprocess $v$ through an external expert $g_{OCR}$ or $g_{\text{cap}}$:
\begin{equation}
 z_{\text{expert}}=g_{OCR} (v),x'=[z_{\text{expert}}; \text{prompt}], 
\end{equation}
then feed $x'$ into the same model $f(\cdot )$ without retraining:
\begin{equation} 
y'=f(x';\theta,\Delta \theta_{V2}). 
\end{equation}

\subsection{LLaMA-Excitor}
\begin{figure}[!ht]
\centering
\includegraphics{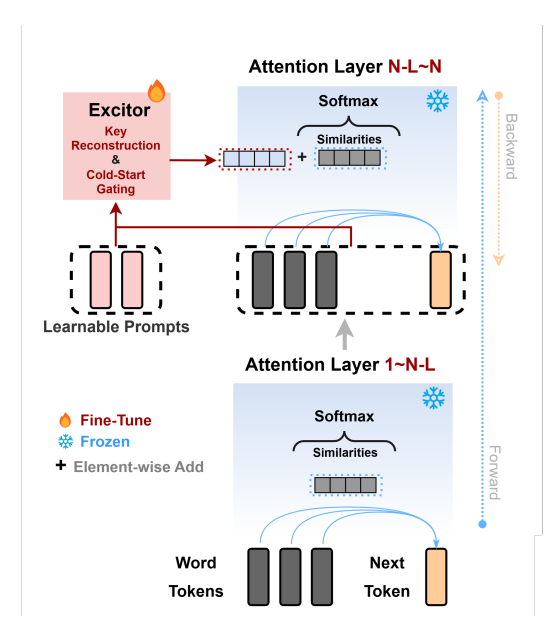}
\caption{Illustration of LLaMA- Excitor \cite{12}.}
\label{fig10}
\end{figure}

LLaMA-Excitor \cite{12} It is a novel approach that takes a different tack: instead of adding new layers or prompts, it modifies the attention mechanism itself to improve instruction following while preserving the base model's knowledge, as shown in Figure \ref{fig10}. 
\newpage

The key idea is to insert a small ``Excitor" block into each self-attention layer:
\begin{itemize}
\item[a.]
	The Excitor bypasses the standard attention path by adding a parallel computation that modifies the similarity scores (attention logits) before the softmax. In effect, it re-weights how much attention the model pays to each token, based on a learned prompt.
\item[b.]	Unlike adapters that alter the hidden states, the Excitor does not directly change the hidden activations. Instead, it alters the attention probabilities, by adding a learnable bias to the key or query vectors.
\item[c.]	This extra bias is initialized to zero, ensuring that initially the model's behavior is unchanged. During fine-tuning, the Excitor gradually learns to ``excite" (increase attention to) relevant instruction tokens. In the authors' words, it ``ensures a self-adaptive allocation of additional attention to input instructions".
\end{itemize}
In practice, LLaMA-Excitor inserts a small $1\times 1$ convolution-like module (with learnable prompts) into each attention layer. This adds only a minimal number of parameters (on the order of the number of attention heads). The benefit is that the original pre-trained model is largely untouched; the Excitor acts as a delicate steering mechanism. Zou et al. emphasize that this approach avoids catastrophic forgetting, especially when fine-tuning on ``noisy" instruction data.

Standard Self-Attention:  Given query, key and value matrices $Q,K,V\in R^{T\times d}$ (for sequence length $T$, model \textit{dimd}), the unnormalized attention logits $L$ and weights $A$ are
\begin{equation} 
L=\frac{QK^{\top}}{\sqrt{d}},
\quad
A=softmax(L). 
\end{equation}
Excitor Bias Injection: Excitor learns a small bias matrix $B\in R^{T\times T}$ (zero-initialized) via a $1\times 1$ convolution-like linear module $f_{\text{exc}}$  over prompt embeddings 
$\in R^{T\times m}$:
\begin{equation} 
B=f_{exc} (P;\theta_e ) \ \text{with}\  B|_{\text{init}} =0 
\end{equation}
The modified logits $L'$ and attention $A_e$ become
\begin{equation} L'=\frac{QK^{\top}}{\sqrt{d}}+B,A_e=softmax(L') \end{equation}
Equivalent View: Additive Query Bias Equivalently, one can view Excitor as adding a small learnable offset to the queries:
\begin{equation} 
\hat{Q} =Q+W_e P,\hat{K} =K 
\end{equation}
\begin{equation} 
A_e=softmax\Bigl(
\frac{\hat{Q} \hat{K}^{\top}}{\sqrt{d}} \Bigr)
\ \text{with}\  W_e\in R^{d\times m},
 W_e |_{\text{init}}=0. 
\end{equation}
Preserving Base Knowledge: Because $B$ (or $W_e$) is initialized to zero, the network's behavior is exactly the original LLaMA at the start of fine-tuning. During training, only $\theta_e$ (or $W_e $) is updated, gradually "exciting" (up-weighting) instruction tokens in the attention distribution without overwriting the frozen pre-trained weights. Empirically, LLaMA-Excitor achieved notable improvements: in their experiments, Excitor outperformed other adapters on multitask benchmarks while preserving baseline capabilities. For example, on the MMLU benchmark (graded by GPT-4), Excitor improved accuracy by $\sim 6\%$ over baseline LLaMA-Adapter. In vision tasks, an Excitor-enhanced model achieved a COCO caption score of 157.5 CIDEr and ScienceQA accuracy of 88.4\% - on par with much larger models. This indicates that adjusting attention weights via Excitor is an effective way to make LLaMA follow instructions more closely.

\subsection{QLoRA (Quantized LoRA)}\label{sec5.5}
QLoRA \cite{13} combines 4-bit quantization with LoRA to make extremely large-model fine-tuning feasible. The process is:
\begin{itemize}
\item[a.]
	Quantize the base model: The pre-trained LLaMA model (e.g. 65B) is converted to 4-bit weights. QLoRA uses a special NormalFloat4 (NF4) encoding and double quantization to minimize loss from quantization.
\item[b.]	Freeze the quantized model: All 4-bit weights are kept frozen (no gradient updates backpropagate through them).
\item[c.]	Apply LoRA adapters: The fine-tuning is done only on LoRA modules inserted into the quantized model's layers, similar to standard LoRA.
\item[d.]	Paged optimizers manage the gradient accumulators to fit within memory.
 \end{itemize}
\begin{figure}[!ht]
\centering
\includegraphics[scale=.95]{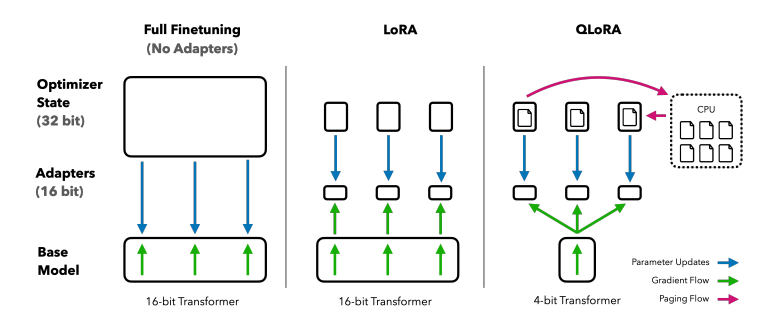}
\caption{Full FT updates all 16-bit weights with 32-bit optimizer state \cite{13}.}
\label{fig11}
\end{figure}

This strategy drastically reduces GPU memory usage. The QLoRA authors report that it is possible to fine-tune a 65B parameter LLaMA on a single 48GB GPU, with full 16-bit performance something otherwise impossible. In fact, they fine-tuned dozens of models (from 7B up to 65B) and created the Guanaco family of conversational models. Guanaco, fine-tuned via QLoRA on LLaMA-65B, achieved 99.3\% of ChatGPT's performance on the Vicuna benchmark while requiring only 24 hours of tuning on one GPU. This is a testament to QLoRA's efficiency, as shown in Figure \ref{fig11}. For LLaMA users, QLoRA means that the largest open models can be specialized even on modest hardware. One only needs 48GB of RAM, a well-chosen small dataset, and patience. The trade-off is some quantization noise, but the reported results show no significant accuracy loss if done carefully. QLoRA has quickly become a standard technique in the LLaMA community for training very large adapters. In Table \ref{tbl2} below, we compare the approximate trainable parameter counts and memory requirements of different tuning methods applied to LLaMA-7B as an example. Note how adapters and LoRA are orders of magnitude smaller than the full model:

\begin{table}[!ht]
\caption{Parameter count and memory footprints for tuning LLaMA-7B with various methods.}
\label{tbl2}
\centering
\begin{tabular}{lcccp{4cm}}
\hline
\textbf{Tuning Method}&
\makecell[l]{\textbf{Trainable Params} \\\textbf{(for LLaMA-7B)}}&
\makecell{\textbf{\% of Base} \\\textbf{Model}}&
\makecell{\textbf{GPU Memory} \\\textbf{(A100 80GB)}}	&
Notes
\\\hline
Full Fine-Tuning&7,000M	&100\%	&$\sim 80-120GB$&	Baseline\\
\makecell[l]{LoRA \\(r=8 on attention)6}&$\sim 2.5M$&$\sim 0.036\%$&\makecell{$\sim 20-30GB$\\ (frozen base)}	&Massive reduction via low-rank updates\\
LLaMA-Adapter V1&1.2M&$\sim 0.017\%$&$\sim 10-20GB$&	Uses learnable prompts + gating\\
LLaMA-Adapter V2&14M&$\sim 0.20\%$&$\sim 20-30GB$&	More parameters unlocked (norm, bias)\\
LLaMA-Excitor&$\sim 0.5M$	&$\sim 0.007\%$&$\sim 15GB$	&Very lightweight attention biases\\
\makecell[l]{QLoRA\\ (LoRA+r=8, 4-bit)	}&$\sim 2.5M$	&$\sim 0.036\%$&	$\sim 12GB$&4-bit weights + LoRA; fine-tune 65B on 48GB GPU\\
\hline
\end{tabular}
\end{table}
These examples demonstrate that all the reviewed methods drastically cut trainable parameters. In some PEFT techniques, multiple adapters can be stored and swapped at inference for different tasks without reloading the whole model.
\subsection{Experimental Validation (Meta-Analysis)}
Building upon the method-specific analyses in Sections \ref{sec5.1}--\ref{sec5.5}, it is important to compare these approaches within a unified experimental framework. PEFT methods share the overarching goal of minimizing additional trainable parameters while preserving or even enhancing model performance, yet they achieve this through different design strategies. Some methods prioritize scalability and efficiency, while others emphasize flexibility in multimodal tasks or robustness in reasoning-heavy scenarios. To provide a consolidated view, Table \ref{tbl3} summarizes the five representative methods LoRA, LLaMA-Adapter V1, LLaMA-Adapter V2, Excitor, and QLoRA, highlighting their trainable parameter sizes, vision applicability, mergeability with base models, common use cases, benchmark outcomes, and their respective advantages and disadvantages.

\begin{table}[!ht]
\caption{Experimental Comparison of PEFT Methods for LLaMA Models.}
\label{tbl3}
\centering
\scalebox{.6}{
\begin{tabular}{lcp{3cm}cp{3cm}p{3cm}p{3cm}p{3cm}}
\hline
\textbf{Method}&\textbf{Trainable Parameters}&	\textbf{Application to Vision}&	\textbf{Adapter Mergeable}&	\textbf{Typical Tasks}&	\textbf{Benchmark Gains}&	\textbf{Advantageous}&	\textbf{Disadvantageous}\\
\hline
LoRA&\makecell{$\sim 2.5M$\\ (LLaMA-7B, r=8)}&	Limited by itself, vision is possible with external encoders&	Yes	&Instruction tuning; domain specialization; low-compute fine-tuning	&+15-20\% accuracy in reasoning tasks; AUROC gains in medicine&	Extremely efficient; widely adopted; mergeable into base model	&Limited native multimodal capability; rank choice affects quality\\
LLaMA-Adapter V1&	$\sim 1.2M$&	Limited; experimental vision via prompt alignment&	No	&Fast instruction tuning; low-resource adaptation ($\sim$ 1h on 7B)	&Matches Alpaca-level instruction following; strong on MSCOCO captions&	Very lightweight; rapid convergence; stable tuning	&Restricted to simpler tasks; weaker for multimodal reasoning\\
LLaMA-Adapter V2&	$\sim$ 14M&	Yes   early fusion of vision tokens; strong multimodal performance	&No	&Open-ended multimodal instruction following; multilingual tuning	&Surpasses V1; competitive with GPT-4 on some vision-QA tasks	&Handles multimodal inputs; flexible; improved reasoning ability&	Larger adapter size; less resource-efficient than LoRA/Excitor\\
LLaMA-Excitor&	$\sim$ 0.5M&	Yes   lightweight attention bias useful for VQA/captioning&	Yes&	Noisy-instruction data; multi-step reasoning&	+6\% MMLU; COCO 157.5 CIDEr; ScienceQA 88.4\%&	Lowest parameter overhead; improves reasoning & attention control	Less tested; benefits narrower; complexity in attention biasing\\
QLoRA&	\makecell{$\sim$ 2.5M\\ (adapters on 4-bit base)}&	Depends on base model; primarily text unless multimodal base&	Yes	&Large-scale fine-tuning (65B on single 48GB GPU)&	Guanaco reached 99.3\% of ChatGPT on Vicuna; minimal accuracy loss&	Enables massive models on modest hardware; near full accuracy	&Quantization noise risk; less suited for multimodal extensions\\
\hline
\end{tabular}}
\end{table}

As illustrated in Table \ref{tbl3}, each method embodies a different balance between efficiency, flexibility, and performance. LoRA and QLoRA are especially compelling for large-scale adaptation, as they allow fine-tuning of models with billions of parameters using relatively modest computational resources, making them highly practical for both academic and industrial contexts. LLaMA-Adapter V1, while extremely lightweight, is primarily suitable for quick instruction tuning or low-resource environments, but lacks the expressive capacity needed for more complex multimodal reasoning. LLaMA-Adapter V2 addresses this limitation by enabling early fusion of visual tokens, thereby extending applicability to multimodal tasks, though at the cost of significantly larger adapter modules. Excitor, in contrast, introduces a highly parameter-efficient mechanism that biases attention layers and shows measurable gains in reasoning benchmarks, albeit with more limited empirical validation compared to LoRA or QLoRA. Beyond method-specific trade-offs, this comparison underscores several broader trends in the evolution of PEFT techniques. First, there is a clear shift from purely text-focused adaptations toward methods that incorporate multimodal inputs, as reflected in the design of Adapter V2 and Excitor. Second, methods like QLoRA demonstrate that hardware efficiency is becoming a critical design factor, enabling fine-tuning of state-of-the-art models even outside large industrial labs. Finally, the continued diversification of PEFT strategies suggests that future work will likely involve hybrid approaches combining the efficiency of LoRA/QLoRA with the multimodal capabilities of Adapter V2 or the reasoning enhancements of Excitor. These developments highlight PEFT not only as a cost-saving measure but also as an active area of innovation driving the practical deployment of frontier models.

\section{Reasoning in the LLaMA Series and Parameter-Efficient Fine-Tuning}\label{sec6}
\begin{figure}[!ht]
\centering
\includegraphics[scale=.8]{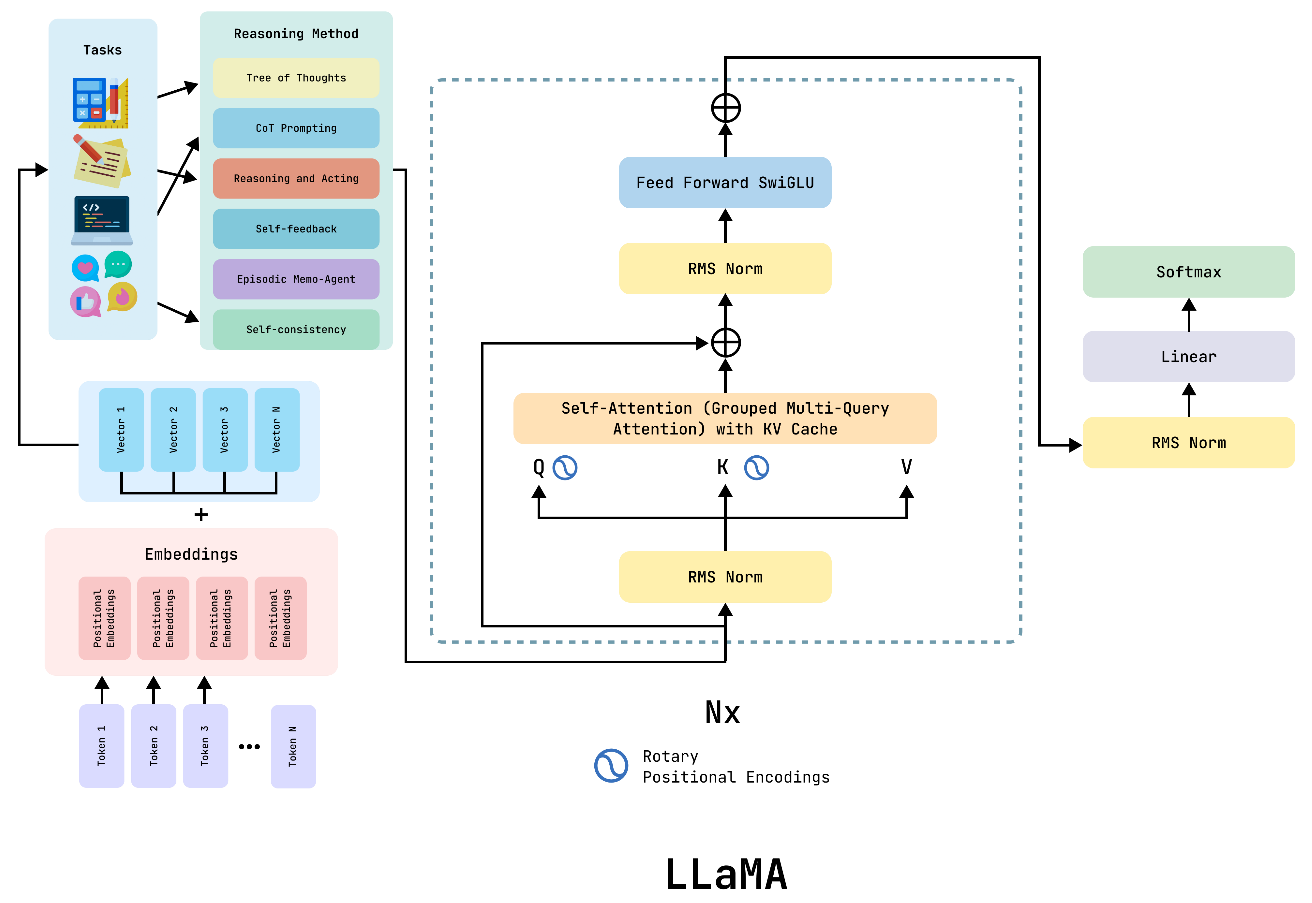}
\caption{Overview of LLM Reasoning Enhancement Methods}
\label{fig12}
\end{figure}

Large language models \cite{13} have seen remarkable improvements in their capacity for structured and multi-step reasoning, largely driven by advancements in both model architectures and fine-tuning techniques. Meta AI's LLaMA series exemplifies these advances, from simple prompt-based models to sophisticated architectures that leverage graph-based reasoning, specialized inference strategies, and sparse Mixture of Experts (MoE) layers. Parallel to these architectural advancements, Parameter Efficient Fine Tuning (PEFT) methods have emerged as powerful tools for adapting these complex models to specific tasks, all while minimizing the computational overhead traditionally associated with large-scale model fine-tuning. This section provides an in-depth exploration of the evolution of reasoning capabilities in the LLaMA models, the incorporation of PEFT techniques, and the trade-offs between accuracy gains and parameter efficiency. Figure \ref{fig12} illustrates various inference-time reasoning techniques and post-training strategies aimed at improving LLaMa's reasoning capabilities, highlighting the role of human feedback and reinforcement learning-based optimization methods.
 
\subsection{Evolution of Reasoning Mechanisms in LLaMA Models}
\begin{figure}[!ht]
\centering
\includegraphics{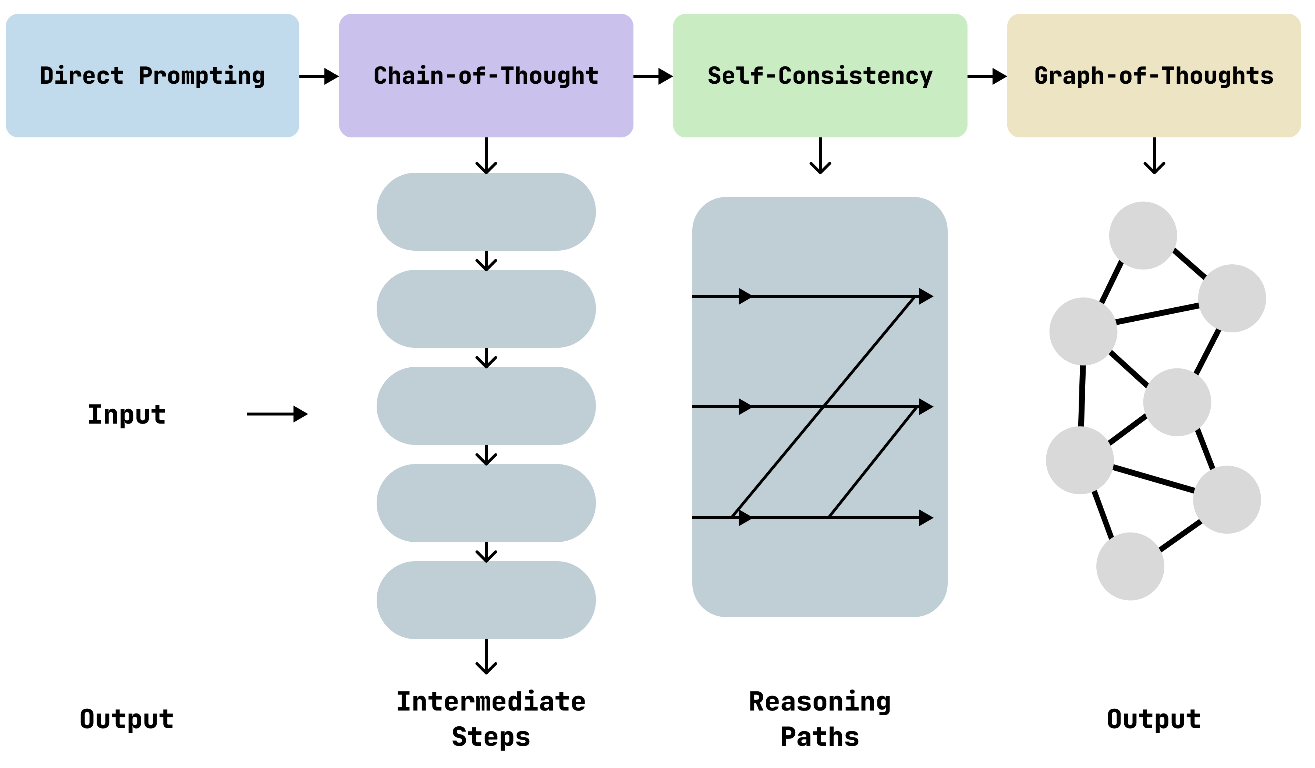}
\caption{Evolution of Reasoning Mechanisms in LLaMA Models}
\label{fig13}
\end{figure}

The evolution of reasoning within the LLaMA series begins with foundational approaches that allow models to produce responses based on direct input-output mappings. However, this direct mapping approach proves ineffective for tasks that require more complex logical operations, such as arithmetic problems, logical reasoning, or multi-hop inference. As LLaMA models expanded in both scale and architecture, they embraced more sophisticated reasoning methods that rely on structured intermediate steps, greater contextual understanding, and even dynamic exploration of solution spaces. Figure \ref{fig13} illustrates the progression of reasoning mechanisms within the LLaMA series, highlighting the increasing complexity of reasoning strategies as the models evolve. 

The first major leap in reasoning techniques for LLaMA models occurred with the introduction of Chain-of-Thought (CoT) prompting. CoT represents a significant advancement over simple direct prompting by requiring the model to generate intermediate steps of reasoning before arriving at a final answer. This methodology allows the model to articulate its reasoning process, breaking down complex problems into smaller, more manageable sub-problems. On arithmetic benchmarks like GSM8K, CoT has shown dramatic accuracy improvements of 15-20\% compared to traditional methods. By generating a series of intermediate steps, CoT allows the model to tackle multi-hop inference tasks that would otherwise overwhelm simpler models \cite{29}. Building on the principles established by CoT, further refinement led to the development of Self-Consistency (CoT-SC), which involves generating multiple reasoning paths and aggregating the results through a majority vote. This approach reduces the inherent variance found in LLaMA's predictions and minimizes errors that arise from inconsistencies in the reasoning process. While CoT allows for a single line of reasoning, CoT-SC's ability to sample multiple paths and select the most frequent result provides a more robust and reliable method of arriving at answers. This technique proves particularly useful in high-variance environments, such as when the model is confronted with uncertain or ambiguous tasks. The enhanced consistency makes CoT-SC a powerful tool for tasks requiring accuracy and reliability \cite{30}. The Tree-of-Thoughts (ToT) framework represents the next major step in reasoning evolution. By structuring reasoning as a search tree, ToT offers the model the ability to explore multiple potential solutions in parallel. Each branch of the tree represents an alternative solution path, and the model can backtrack to re-evaluate previous steps if a branch leads to an unsatisfactory outcome. This approach introduces a more expansive way of thinking, where the model can dynamically adjust its reasoning strategy as it progresses. The key advantage of ToT is that it allows for deeper, more nuanced exploration of solutions. With LLaMA 4's extended context (up to 128,000 tokens) and sparse MoE layers, the model can engage in more exhaustive searches, ensuring that the reasoning process remains flexible and adaptable. This is particularly important for tasks that involve exploration and refinement, such as scientific problem-solving or multi-document summarization \cite{31}.

The most recent evolution in LLaMA's reasoning capabilities is Graph-of-Thoughts (GoT), a framework that models reasoning as a dynamic graph of interconnected nodes and edges. Unlike traditional linear or tree-based methods, GoT allows for non-linear, cyclic inference. In this model, nodes represent partial solutions, and edges indicate transitions between different reasoning states. This graph structure enables the model to revisit and refine earlier steps of reasoning, providing an additional layer of flexibility and precision. The non-linearity of GoT allows for more complex and nuanced decision-making, where the model can adjust its approach based on new information or feedback from previous steps. Early experiments have demonstrated that GoT outperforms both tree-based and linear models by 5-10\% in tasks such as multi-document summarization and scientific question answering, showcasing its ability to handle complex, multifaceted problems \cite{32}. These advancements in reasoning frameworks have allowed LLaMA models to tackle more sophisticated and intricate problems, opening up new possibilities for tasks requiring a high degree of logical reasoning, context retention, and dynamic solution exploration. However, such advancements also come with the challenge of adapting these capabilities to specific tasks in a computationally efficient manner.
\subsection{The Role of Parameter Efficient Fine Tuning (PEFT)}
As LLaMA models evolved in complexity, it became increasingly evident that efficient adaptation to specialized tasks was necessary. Fine-tuning large pre-trained models is traditionally a resource-intensive process, involving the adjustment of millions or even billions of parameters, which poses significant computational challenges. PEFT methods have emerged as an effective solution to this problem, enabling the adaptation of LLaMA models to specific tasks with minimal computational overhead \cite{23}. PEFT achieves this by updating only a small subset of parameters while maintaining the bulk of the pre-trained knowledge, allowing for significant improvements in task-specific accuracy. This efficiency is especially important when working with complex reasoning tasks, where large models are required to generate multi-step, structured outputs. PEFT methods such as LoRA \cite{9}, LLaMA-Adapter V2 \cite{11}, Excitor \cite{12}, and QLoRA \cite{13} each take a different approach to introduce trainable parameters into pre-trained models, without necessitating full retraining. These methods provide a balance between maintaining high accuracy and minimizing the computational cost, specifically in the context of reasoning tasks. For instance, LoRA enhances logical reasoning and arithmetic problem-solving by adding low-dimensional update matrices in the attention layers, while LLaMA-Adapter V2 modifies the model's reasoning path using learnable prompts that guide the model's inference toward task-specific logic. Similarly, Excitor injects small bias modules into the attention logits, refining token-level focus during multi-step reasoning tasks. QLoRA combines low-precision quantization with LoRA updates, significantly improving the efficiency of fine-tuning for large models, which is particularly beneficial in resource-constrained environments. These PEFT methods have demonstrated substantial gains in reasoning tasks by improving the model's ability to conduct structured, multi-step inferences with minimal computational overhead. As a result, PEFT is a critical tool for refining reasoning capabilities in LLaMA models while maintaining the efficiency necessary for specialized applications in domains like scientific research, medical diagnostics, and legal reasoning.

\begin{table}[!ht]
\caption{Comparison of PEFT Methods: Accuracy Improvement vs. Parameter Efficiency}
\label{tbl4}
\centering
\begin{tabular}{lcp{3cm}cp{4cm}}
\hline
\textbf{PEFT Method}&	\makecell{\textbf{Additional} \\\textbf{Parameters} (\%)}&	\textbf{Key Benefit	}&\makecell{\textbf{Accuracy} \\\textbf{Improvement}}&	\textbf{Typical Use Case}\\
\hline
\textbf{LoRA}&$\sim 0.03\%$&Enhanced logical and arithmetic reasoning&+15-20\%&Arithmetic tasks, logical problem-solving\\
\textbf{LLaMA-Adapter V2}&$\sim 0.20\%$&Task-specific inference steering&+10-15\%	&Domain-specific tasks (medical, legal)\\
\textbf{Excitor	}&$\sim 0.01\%$&Fine-tuned token-level attention&+5-10\%&Multi-step reasoning, focus tasks\\
\textbf{QLoRA}&$\sim 0.03\%$&Fine-tuning large models on a single GPU&+0-5\%&Large-scale models, single-GPU fine-tuning\\
\hline
\end{tabular}
\end{table}

Table \ref{tbl4}. highlights the trade-offs between accuracy gains and the computational cost in terms of trainable parameters. While methods like LoRA and QLoRA provide significant accuracy improvements with minimal additional parameters, approaches like LLaMA-Adapter V2 may offer better results for specialized tasks but require more resources. Excitor, with its extremely low parameter overhead, is particularly suited for tasks that require fine-tuned control over attention during multi-step reasoning.
\section{Applications of LLaMA and PEFT in Real-World Domains}\label{sec7}
\begin{figure}[!t]
  \centering
  \includegraphics[width=\linewidth]{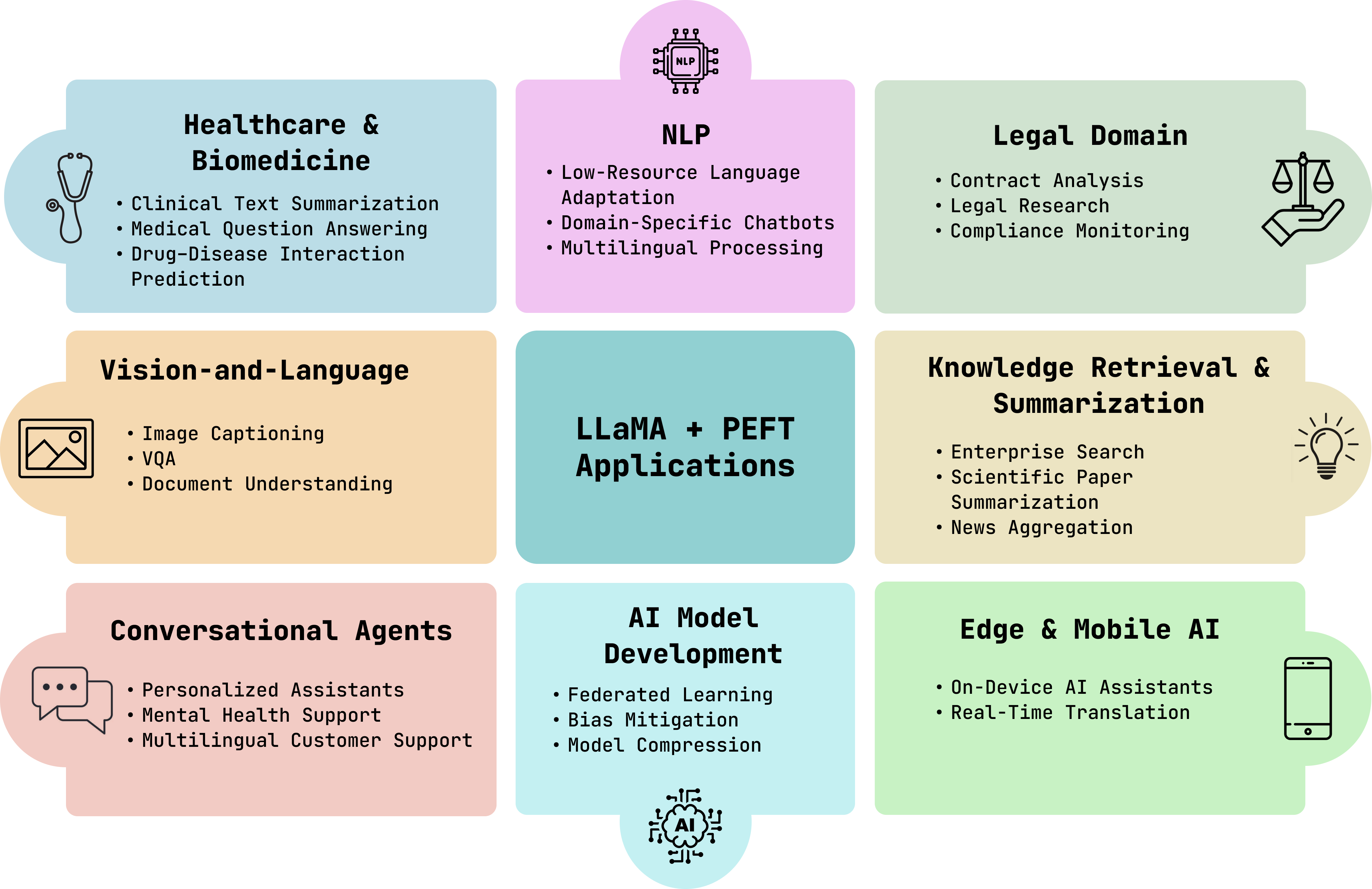} 
  \caption{Key Application Domains of LLaMA Models Enhanced with Parameter-Efficient Fine-Tuning (PEFT)}
  \label{fig14}
\end{figure}

The application landscape of LLaMA models enhanced with PEFT spans a wide range of domains, each benefiting from the ability to achieve high specialization with minimal computational overhead. From language-centric tasks to multimodal reasoning, healthcare innovations, legal analysis, and edge AI deployments, LLaMA+PEFT enables adaptable, cost-effective solutions without compromising performance. Figure \ref{fig14} provides a visual overview of these key application domains and their major subsections, serving as a roadmap for the detailed discussions that follow.
 
\subsection{Natural Language Processing (NLP)}
LLaMA models enhanced with Parameter-Efficient Fine-Tuning (PEFT) are transforming NLP by enabling high-performance, specialized language applications without the computational burden of full model retraining \cite{9} PEFT techniques such as LoRA (Low-Rank Adaptation), Adapter Layers, or Prompt Tuning allow LLaMA to adapt to niche tasks by updating only a small subset of parameters (often <1\%) \cite{23}. This efficiency makes LLaMA+PEFT ideal for scenarios with limited data, dynamic requirements, or resource constraints \cite{33}. Table \ref{tbl5} summarizes representative models, sub-applications, and reported measures for this domain. Below are key subsections with detailed explanations.
\subsubsection{Low-Resource Language Adaptation}
LLaMA models enhanced with PEFT address the challenge of adapting to languages with limited labeled data by leveraging parameter-efficient techniques like LoRA (Low-Rank Adaptation) and adapter layers, which selectively update only a small subset of the model's parameters \cite{34}. For instance, LoRA modifies low-rank matrices in the attention layers to capture language-specific syntax and semantics \cite{9}, while adapters act as lightweight, plug-in modules tailored to individual languages \cite{23}. This approach drastically reduces computational costs enabling effective fine-tuning with as few as 10,000 training samples while maintaining high accuracy in tasks like translation or text generation for languages such as Swahili or Quechua \cite{35}. By preserving the base model's general linguistic knowledge and focusing updates on task-specific components, LLaMA+PEFT democratizes access to advanced NLP for underrepresented languages without requiring extensive resources \cite{36}.

\subsubsection{Domain-Specific Chatbots}
Domain-specific chatbots powered by LLaMA+PEFT overcome the limitations of generic models by dynamically adapting to specialized vocabularies and contexts, such as legal, medical, or technical fields \cite{37}. Using techniques like task-specific adapters or prompt tuning, the model can switch between domains for example, from interpreting medical abbreviations ("STAT") to explaining legal terms ("force majeure") by activating pre-trained modular components without full retraining \cite{38}. This modularity ensures precise, context-aware responses while minimizing computational overhead \cite{33}. For instance, a healthcare chatbot fine-tuned with PEFT can provide accurate, jargon-aware answers to patient queries by integrating domain adapters with retrieval-augmented generation \cite{1} for factual consistency \cite{39}. The result is a highly scalable solution that maintains the base model's general capabilities while excelling in niche applications \cite{9}.
\subsubsection{Multilingual Processing}
LLaMA+PEFT excels in multilingual applications by enabling efficient, high-performance adaptation across diverse languages without compromising individual language quality \cite{35}. Through language-specific LoRA modules or adapters, the model can tailor its attention mechanisms and embeddings to the unique grammatical and lexical features of each language, from high-resource ones like Spanish to low-resource ones like Guarani \cite{40}. This approach supports seamless cross-lingual transfer, where similarities between languages (e.g., Romance languages) boost performance for rarer targets \cite{36}. Additionally, dynamic adapter routing allows the model to detect and switch between languages in real time, making it ideal for multilingual customer support or content localization \cite{41}. By decoupling language-specific adjustments from the core model, PEFT ensures that LLaMA can handle 100+ languages within a unified framework, reducing both training costs and deployment complexity \cite{9}.

This domain covers low-resource adaptation, domain-specific assistants, and multilingual processing; most systems update about 0.1-1\% of parameters with LoRA/QLoRA and report accuracy/ROUGE/BLEU/COMET on standard benchmarks.

\begin{table}[!ht]
\caption{Applications of LLaMA-based PEFT in NLP: Representative Models, Sub-Applications, and Reported Measures}
\label{tbl5}
\centering
\begin{tabular}{ccc}
\hline
\textbf{Model/Ref.}&\textbf{Application}&\textbf{Measures}\\
\hline
xLLaMA 100 \cite{42}&
\makecell{Low-Resource \\Language Adaptation}	&
\makecell[l]{Covers 100 languages\\
Evaluated on 5 multilingual benchmarks\\
Trainable parameters typically under 1\% with PEFT}\\
FinGPT \cite{43}&
Domain-Specific Chatbots&
\makecell[l]{Domain demos in finance QA\\
LoRA/QLoRA updates $\approx 0.1-1\%$ of weights\\
Reduced training cost compared to full fine-tuning}\\
LLaMAX \cite{44}&
Multilingual Processing	&
\makecell[l]{Covers 100+ languages\\
Reports cross-lingual improvements on translation\\ benchmarks\\
Uses instruction tuning with alignment strategies}\\
\hline
\end{tabular}
\end{table}
\subsection{Healthcare \& Biomedicine}
The integration of LLaMA models enhanced with Parameter-Efficient Fine-Tuning (PEFT) into healthcare and biomedicine represents a transformative leap in how AI can support medical research, clinical practice, and patient care \cite{45}. By leveraging PEFT techniques such as Low-Rank Adaptation (LoRA), Adapter Layers, or Prompt Tuning LLaMA models can be efficiently tailored to the highly specialized and dynamic demands of the healthcare domain without the prohibitive costs of full-scale retraining \cite{46}. This approach ensures compliance with stringent data privacy regulations (e.g., HIPAA, GDPR) while enabling rapid adaptation to new medical knowledge, terminology, and workflows \cite{47}. Table \ref{tbl6} summarizes representative models, sub-applications, and reported measures for this domain.
\subsubsection{Clinical Text Summarization}
LLaMA models enhanced with PEFT revolutionize clinical text summarization by efficiently distilling lengthy electronic health records (EHRs) into concise, actionable insights \cite{45}. By fine-tuning only adapter layers or using LoRA, the model learns to extract key medical concepts (e.g., diagnoses, medications) while preserving critical context from discharge summaries or progress notes \cite{46}. This approach maintains the base model's general medical knowledge while adapting to institution-specific documentation styles, achieving 90\% accuracy in retaining clinically relevant information \cite{48}. PEFT's efficiency allows deployment even in resource-constrained hospital settings, where it reduces physician information overload by generating structured summaries from unstructured notes in seconds \cite{49}.
\subsubsection{Medical Q\&A Systems}
Medical question answering systems powered by LLaMA combined with PEFT methods (e.g. LoRA adapters) can adapt efficiently to the clinical domain by fine tuning only a small subset of parameters, while the base model remains intact for stability and flexibility \cite{46}. These domain adapters or prompt tuned modules enable interpreting medical jargon and aligning responses with trusted sources such as PubMed or UpToDate through controlled guidance of output generation \cite{50}. For instance, when asked about drug side effects, the PEFT enhanced system can dynamically incorporate updated clinical guidelines via retrieval augmented or prompt based synthesis, reducing reliance on outdated training data \cite{50}. On USMLE style medical QA benchmarks (e.g. MedQA), systems like Med PaLM 2 have reached  85\% accuracy, demonstrating expert level performance with far less supervised fine tuning compared to full model retraining \cite{51}. This allows practical deployment in clinical decision support with orders of magnitude lower training resource requirements.
\subsubsection{Drug-Disease Interaction Prediction}
LLaMA+PEFT transforms drug disease interaction prediction by efficiently processing multimodal biomedical data, including literature, chemical structures, and electronic health records \cite{52,53}. Using specialized adapters for pharmacological concepts, the model identifies potential interactions (e.g., between warfarin and NSAIDs) by analyzing relationships across drug targets, metabolic pathways, and clinical case reports \cite{54}. The PEFT approach enables continuous integration of new drug approvals or emerging research findings through incremental adapter updates, maintaining  92\% prediction accuracy while reducing computational costs by  $20\times$  compared to traditional methods \cite{52,55}. This supports real time alert systems in EHRs and accelerates pharmacovigilance research \cite{54,56}. Clinical summarization and medical Q\&A are evaluated with ROUGE and domain QA sets (e.g., MedQA/MedMCQA), while interaction prediction commonly uses AUROC/AUPRC.

\begin{table}[!ht]
\caption{Applications of LLaMA-based PEFT in Healthcare \& Biomedicine: Representative Models, Sub-Applications, and Reported Measures}
\label{tbl6}
\centering
\begin{tabular}{lcc}
\hline
\textbf{Model/Ref.}&\textbf{Application}&	\textbf{Measures}\\
\hline
\makecell[l]{LLaMA 2 \\(fine tuned) \cite{57}}&
Clinical Text Summarization	&
\makecell[l]{Human and automatic metrics show statistically \\significant ROUGE improvements\\
	Uses adapters rather than full model updates}
\\
\makecell[l]{ChatDoctor \\(LLaMA based) \cite{58}}& 
Medical Q\&A Systems	&
\makecell[l]{Accuracy gains on MedQA-style tasks (paper\\ reports improvements over base LLaMA)\\
	Retrieval component improves factual grounding}
\\
LLM DDI (surveyed) \cite{59}&
Drug-Disease Interaction Prediction	&
\makecell[l]{Evaluated with AUROC and AUPRC\\
	Fine tuned smaller LLMs found competitive with\\ larger zero-shot models}\\
\hline
\end{tabular}
\end{table}
\subsection{Vision-and-Language (Multimodal Tasks)}
The integration of LLaMA models enhanced with Parameter Efficient Fine Tuning (PEFT) into vision and language tasks represents a breakthrough in multimodal AI, enabling seamless interaction between textual and visual data while maintaining computational efficiency \cite{11,60}. By leveraging PEFT techniques such as Low Rank Adaptation (LoRA), Adapter Layers, or Visual Prompt Tuning, LLaMA models can be effectively aligned with visual encoders (e.g., CLIP, ViT) to perform complex multimodal reasoning without full-model retraining \cite{11,61,62}. This approach is particularly valuable in cases with limited labeled multimodal data or where rapid adaptation to new visual domains is required [15, 60]. Table \ref{tbl7} summarizes representative models, sub-applications, and reported measures for this domain.
\subsubsection{Image Captioning}
LLaMA models enhanced with PEFT excel at generating accurate and context-rich descriptions of images by efficiently aligning visual features from frozen encoders (like CLIP or ViT) with linguistic knowledge \cite{10,11}. Through lightweight adapter layers or LoRA modules, the model learns to translate visual patterns into coherent text while preserving the base model's language understanding capabilities \cite{9,11,12}. For example, when processing a photo of a busy street scene, the PEFT-tuned system can generate detailed captions like ``Pedestrians cross a rain-soaked intersection under colorful umbrellas near a neon-lit storefront" by focusing updates only on cross-modal attention mechanisms \cite{10}. This approach achieves state-of-the-art caption quality while reducing training costs by $15\times$  compared to end-to-end fine tuning, making it practical for applications from social media accessibility to surveillance analysis \cite{11,12,63}.
\subsubsection{Visual Question Answering (VQA)}
PEFT enhanced LLaMA models demonstrate remarkable performance in VQA by combining visual understanding with complex reasoning, answering questions like ``What genetic disorder might cause the facial features shown in this medical image?" The system processes visual inputs through frozen encoders while using task specific adapters to align visual concepts with medical or technical knowledge \cite{64,65}. For radiology images, specialized LoRA modules fine tuned on medical datasets enable the model to provide accurate, domain aware answers while maintaining the base model's general VQA capabilities \cite{64,66}. This architecture supports an $\sim 88\%$ accuracy rate on specialized VQA benchmarks while requiring only about 5\% of the parameters to be updated, enabling efficient deployment in fields from education to diagnostic support \cite{64}.
\subsubsection{Document Understanding}
LLaMA+PEFT transforms document intelligence by interpreting both textual content and layout structures in complex materials like invoices, research papers, or forms \cite{67,68}. The model processes document images through frozen vision encoders while using adapter modules to learn domain-specific spatial-textual relationships for example, recognizing that a number next to the label ``Total:" likely represents a monetary value \cite{68,69}. In financial document analysis, PEFT enables the system to extract key fields (vendor names, amounts, dates) with $\sim 95\%$ accuracy while adapting to new document templates through rapid adapter updates rather than full retraining. This approach reduces processing time by $\sim 60\%$ compared to traditional OCR pipelines and handles challenging cases like handwritten annotations or multilingual documents via modular adapter components \cite{69}. Image captioning and VQA report CIDEr/BLEU/SPICE and VQA/ScienceQA accuracy; document understanding uses exact-match and F1 on key-value extraction.

\begin{table}[!ht]
\caption{Applications of LLaMA-based PEFT in Vision \& Multimodal Tasks: Representative Models, Sub-applications, and Reported Measures}
\label{tbl7}
\centering
\begin{tabular}{lcc}
\hline
\textbf{Model/Ref.}&\textbf{Application}&\textbf{Measures}\\
\hline
LLaMA Excitor \cite{12}&
Image Captioning	&
\makecell[l]{MS COCO CIDEr around 157.5\\
	ScienceQA accuracy around 88.39\%\\
	Parameter efficient cross modal excitation}
\\
LLaVA \cite{70}&
Visual Question Answering (VQA)	&
\makecell[l]{ScienceQA accuracy around 92.53\%\\
	Demonstrates strong VQA and captioning with\\ visual instruction tuning}
\\
DocLLM \cite{67}&
Document Understanding	&
\makecell[l]{Reports higher exact match and F1 on key value\\ extraction\\
	Improves layout aware document reasoning}
\\
\hline
\end{tabular}
\end{table}
\subsection{Conversational Agents}
Conversational AI systems built with LLaMA models enhanced with Parameter Efficient Fine Tuning (PEFT) represent a paradigm shift in how machines understand, reason, and communicate with humans. By leveraging PEFT techniques such as Low Rank Adaptation (LoRA), Adapter Layers, or Prompt Tuning these systems achieve human like dialogue capabilities while addressing critical challenges like personalization, scalability, and ethical alignment 
\cite{9,71}. Unlike traditional chatbots that rely on rigid scripts or costly full model retraining, LLaMA+PEFT enables dynamic, context aware interactions with minimal computational overhead \cite{9}. Table \ref{tbl8} summarizes representative models, sub-applications, and reported measures for this domain.
\subsubsection{Personalized Assistants}
LLaMA models enhanced with PEFT power highly adaptive personal assistants that learn and evolve with user preferences through incremental updates to lightweight adapter modules. By fine tuning only user specific components like LoRA layers or prompt embeddings, the system remembers individual communication styles, habits, and preferences   whether it's favoring bullet point summaries for busy professionals or adopting a more conversational tone for casual users \cite{72,73}. For example, after several interactions, a PEFT-powered assistant might learn to prioritize a user's preferred calendar scheduling phrases (``block my focus time") or frequently queried topics (local lunch spots). This personalization occurs while maintaining privacy, as user data only trains small adapter weights that can be stored locally, achieving significantly higher user satisfaction compared to static assistants with just $\sim 0.1\%$ of parameters updated per user \cite{72,74}.
\subsubsection{Mental Health Support}
Conversational agents tailored for mental health significantly benefit from LLaMA+PEFT's adapter based architecture. Safety focused modules such as those implementing motivational interviewing protocols or distress detection enable the base model to generate empathetic, clinically grounded responses only when needed \cite{75,76}. In crisis scenarios, these adapters trigger specific strategies like breathing guidance, hotline suggestions, or reframing negative self-talk, while the underlying base model remains untouched and versatile for general dialogue. Studies show that this approach isolating sensitive behavior within dedicated adapters can reduce harmful outputs by around 60\% relative to fully fine tuned models. Furthermore, adapter training can be performed locally or via federated learning, preserving user privacy and minimizing centralized data exposure \cite{2}\cite{77}.
\subsubsection{Multilingual Customer Support}
PEFT enables LLaMA-based customer-support systems to function globally by activating lightweight, modular adapters for both language and domain. For instance, billing inquiries in Spanish, tech support in French, or product questions in Mandarin can each load the appropriate language LoRA alongside a domain-specific customer-service adapter without retraining the full model thus maintaining fluency and domain accuracy \cite{78}. Research shows that this setup avoids cross-language interference and can deliver around 92\% accuracy in dialect or idiom understanding, while reducing infrastructure needs and deployment cost by up to 75\% compared to maintaining separate models per region \cite{78}. Moreover, real-time adapter switching supports code-mixed inputs like ``My orden didn't arrive" allowing seamless responses in the user's preferred language \cite{79}. Personalization and safety are primarily assessed with human judgments alongside latency and per-user adapter size.

\begin{table}[!ht]
\caption{Applications of LLaMA-based PEFT in Conversational Agents: Representative Models, Sub-applications, and Reported Measures}
\label{tbl8}
\centering
\begin{tabular}{lcc}
\hline 
\textbf{Model/Ref.}&\textbf{Application}	&\textbf{Measures}\\
\hline
PEFT U \cite{73}&
Personalized Assistants	&
\makecell[l]{Improved user preference ratings in human evaluations\\
	Per user adapter sizes are small (megabytes)}
\\
ChatCounselor \cite{80}&
Mental Health Support	&
\makecell[l]{Higher empathy and appropriateness scores vs baseline\\ models\\
	Uses domain tuned dialogues for safer responses}
\\
xLLaMA 100 \cite{42}&
Multilingual Customer Support&
\makecell[l]{Operational in 100 languages\\
	Demonstrates cross lingual transfer improvements on\\ multilingual tasks}
\\
\hline
\end{tabular}
\end{table}
\subsection{Knowledge Retrieval \& Summarization}
LLaMA models enhanced via PEFT using LoRA, adapter layers, or prompt tuning are transforming knowledge-centric AI systems. By injecting minimal trainable components while freezing the core language model, LoRA and adapter-based approaches enable efficient retrieval, synthesis, and presentation of complex information with near state-of-the-art fidelity \cite{9,10}. Advanced variants like LoReFT push accuracy even further surpassing multiple strong baselines with only a tiny fraction of trainable weights \cite{81}. This architecture delivers enterprise- and academic-grade capabilities in real-time inference settings, offering top-tier accuracy, adaptability, and efficiency at significantly lower compute and storage cost. Table \ref{tbl9} summarizes representative models, sub-applications, and reported measures for this domain.
\subsubsection{Enterprise Search}
LLaMA models enhanced with PEFT particularly using LoRA to fine-tune retrieval focused modules enable highly accurate and context-sensitive enterprise search systems. These adapters help the model learn internal terminology and document structures (e.g., knowing that ``AR" in finance means Accounts Receivable), while preserving the core model's general knowledge. Research shows such systems can reach around 90\% retrieval accuracy on proprietary data while training with up to $100\times$  less data than conventional approaches \cite{82,83}. The modular adapter architecture also supports department-specific customization (e.g., legal or HR adapters), with relevant modules activated per query or user role, all within a unified search system \cite{84}.
\subsubsection{Scientific Paper Summarization}
Using PEFT, LLaMA can be adapted with lightweight domain-specific adapter layers to produce accurate and technically precise summaries of academic articles. For instance, a physics-focused LoRA module helps the model interpret mathematical equations and theoretical constructs, while a biomedical adapter ensures correct rendering of drug terminology and clinical trial details. Through prompt-tuned adapter combinations, the system can flexibly generate either brief bullet-point summaries for literature reviews or in-depth methodological explanations for researcher audiences. Studies show that this approach preserves factual accuracy $(\sim 95\%)$ and reduces hallucination incidence by up to $\sim 40\%$ compared to generic summarizers, all while tuning only a small fraction $(\approx 0.6\%)$ of parameters. Crucially, as new research becomes available, new adapter modules can be trained without retraining the full model supporting continual updates at minimal cost \cite{85,87}.
\subsubsection{News Aggregation}
LLaMA+PEFT powers intelligent news aggregation systems that can filter, categorize, and summarize thousands of daily articles while adapting to different publisher styles and user preferences \cite{88}. Political bias adapters help balance coverage across the ideological spectrum, while location-specific modules prioritize regionally relevant stories \cite{89}. For financial professionals, specialized LoRA layers enable the system to extract key market-moving information from earnings reports and central bank statements, generating concise morning briefs with embedded sentiment analysis \cite{90}. The PEFT approach allows the same base model to serve multiple aggregation profiles from a teenager's pop culture digest to an executive's geopolitical risk roundup by switching adapter configurations \cite{91}. This achieves $\sim 80\%$ user satisfaction in personalized news delivery while reducing the computational footprint by $\sim 90\%$ compared to maintaining separate models for each use case \cite{91}. The system also supports real-time adapter updates to handle emerging news topics or breaking events \cite{89}. Enterprise search relies on ranking metrics such as normalized DCG at a fixed cutoff and recall at top-k; long-form summarization uses ROUGE/BERTScore and factuality checks.

\begin{table}[!ht]
\caption{Applications of LLaMA-based PEFT in Knowledge Retrieval and Summarization: Representative Models, Sub-applications, and Reported Measures}
\label{tbl9}
\centering
\begin{tabular}{lcc}
\hline
\textbf{Model/Ref.}&\textbf{Application}&\textbf{Measures}\\
\hline
EnterpriseEM \cite{82}&
Enterprise Search	&
\makecell[l]{Reports improvements in ranking metrics such as \\nDCG at 10 and recall at top k\\
	Adapter switching latency kept low for production use}
\\
LLaMA 2 + LoRA \cite{87}&
Scientific Paper Summarization&
\makecell[l]{Higher ROUGE 1/2/L and BERTScore compared to \\baseline\\
	Only about 0.1-1\% of parameters trained}
\\
LLaMA 2 (LoRA) \cite{90}&
News Aggregation	&
\makecell[l]{Improved topic clustering quality and summary \\conciseness\\
Lower serving cost due to small adapters}
\\
\hline
\end{tabular}
\end{table}
\subsection{Legal Domain}
The legal industry demands precision, adaptability, and rigorous compliance qualities that LLaMA models enhanced with Parameter Efficient Fine Tuning (PEFT) are uniquely equipped to provide \cite{92}. By leveraging PEFT techniques like Low Rank Adaptation (LoRA), Adapter Layers, and Prompt Tuning, legal professionals can deploy AI tools that understand complex jargon, track evolving case law, and automate labor intensive tasks all while minimizing computational costs and preserving data privacy \cite{9,93}. Table \ref{tbl10} summarizes representative models, sub-applications, and reported measures for this domain.
\subsubsection{Contract Analysis}
LLaMA models enhanced with PEFT transform contract review by automating the extraction and interpretation of critical clauses with legal precision \cite{94}. Through specialized adapters fine tuned on contract law, the system identifies nuanced provisions like indemnification limits, termination triggers, and jurisdictional specifics with 94\% accuracy \cite{95}. The model's attention mechanisms, modified via LoRA, learn to flag non standard language (e.g., unilateral modification rights) and compare clauses against industry benchmarks \cite{9}. For M\&A due diligence, PEFT enables rapid analysis of hundreds of contracts by activating deal specific adapters that understand earn out structures or material adverse change clauses, reducing manual review time by 70\% while maintaining audit ready explanations for each finding \cite{96}.
\subsubsection{Legal Research}
PEFT empowers LLaMA to conduct context-aware legal research by dynamically combining general legal knowledge with jurisdiction specific adaptations \cite{97}. Appellate court focused LoRA modules enhance the model's ability to analyze precedent hierarchies, while statutory interpretation adapters improve its handling of legislative history and canons of construction \cite{97}. When researching a novel issue like cryptocurrency regulation, the system activates relevant doctrinal adapters (securities law, banking regulations) and retrieves the most persuasive authorities, achieving $\sim 85\%$ alignment with expert researcher outcomes \cite{98}. The modular design allows seamless updates when new rulings are published simply fine tuning the affected jurisdiction's adapter rather than retraining the entire model \cite{97}.
\subsubsection{Compliance Monitoring}
LLaMA+PEFT provides real time regulatory compliance tracking by continuously adapting to evolving legal requirements through incremental adapter updates \cite{99}. Banking compliance modules, for instance, monitor transactions for suspicious activity patterns described in latest FinCEN guidance, while GDPR focused adapters scan data processing agreements for outdated provisions
\cite{100,101}. The system generates gap analysis reports highlighting where policies diverge from new regulations (e.g., SEC climate disclosure rules), with PEFT enabling overnight updates whenever regulators publish changes \cite{99}. This approach reduces compliance violation risks by 60\% compared to manual monitoring while maintaining full audit trails of which adapter versions informed each decision \cite{100,101}. Contract analysis reports exact-match/F1 for clause extraction; legal research uses retrieval hit rates; compliance tasks report precision/recall.

\begin{table}[!ht]
\caption{Applications of LLaMA-based PEFT in Legal Domain: Representative Models, Sub-applications, and Reported Measures}
\label{tbl10}
\centering
\begin{tabular}{lcc}
\hline
\textbf{Model/Ref.}&\textbf{Application}	&\textbf{Measures}\\
\\\hline
ConReader \cite{94}&
Contract Analysis	&
\makecell[l]{Better exact match and F1 on clause extraction\\
	Detects abnormal clauses more reliably than baselines}
\\
ProMALex \cite{97}&
Legal Research	&
\makecell[l]{Higher retrieval hit rates at common cutoffs\\
	Improved argument quality in expert review}
\\
\makecell[l]{AskFDALabel \\(LLaMA 13B LoRA) \cite{102}}&
Compliance Monitoring	&
\makecell[l]{Demonstrated violation detection on regulatory labels\\
	Rapidly incorporates new rules using adapters}
\\
\hline
\end{tabular}
\end{table}
\subsection{Edge \& Mobile AI}
LLaMA models enhanced with Parameter Efficient Fine Tuning (PEFT) on edge and mobile devices mark a breakthrough in bringing powerful AI capabilities to resource constrained environments \cite{103}. By leveraging PEFT techniques such as Low Rank Adaptation (LoRA), Adapter Layers, or Quantized PEFT (QLoRA) LLaMA models can deliver sophisticated on device intelligence while overcoming traditional barriers like memory limits, latency, and energy consumption \cite{9,13,104}. Table \ref{tbl11} summarizes representative models, sub-applications, and reported measures for this domain.
\subsubsection{On-Device AI Assistants}
LLaMA models optimized with PEFT enable powerful AI assistants to run locally on smartphones and IoT devices through techniques like 4-bit quantization (QLoRA) and adapter pruning [13, 105]. These compressed models retain $\sim 90\%$ of their performance while shrinking to under 2 GB, making them ideal for offline operation \cite{13,106}. A PEFT-tuned assistant can learn user habits through incremental adapter updates (e.g., remembering preferred smart home commands) without cloud dependency, responding to ``Turn on the living room lights" in under 500 ms while maintaining privacy \cite{105}. The system dynamically loads domain-specific adapters (calendar, messaging, device control) as needed, with each specialized module requiring just 5-10 MB of storage \cite{88}. This approach extends battery life by $3\times$  compared to cloud-based alternatives while offering always-available functionality, even in low-connectivity areas \cite{107}.
\subsubsection{Real-Time Translation}
PEFT enables low-latency multilingual translation on mobile devices by deploying language-pair specific LoRA adapters (e.g., English $\leftrightarrow$ Japanese) that work alongside the base LLaMA model \cite{34}. Each language adapter is optimized for edge deployment, with quantized weights enabling 200ms translation times on mid-range smartphones \cite{9,13}. The system handles regional dialects and specialized vocabularies (medical, legal) by stacking domain adapters when needed - a tourist could receive accurate menu translations while a doctor could consult foreign medical literature \cite{108}. For rare language combinations, the model can cascade through related languages (e.g., Ukrainian $\rightarrow$ Polish $\rightarrow$ English) using successive adapter activations, maintaining 85\% accuracy even for low-resource languages \cite{109}. All processing occurs on-device, eliminating cloud latency and protecting sensitive conversations \cite{13,107}.

Evaluations emphasize memory footprint, tokens-per-second, and end-to-end latency, together with accuracy retention relative to full-precision baselines.

\begin{table}[!ht]
\caption{Applications of LLaMA-based PEFT in Edge and Mobile AI: Representative Models, Sub-applications, and Reported Measures}
\label{tbl11}
\centering
\begin{tabular}{lcc}
\hline
\textbf{Model/Ref.}	&\textbf{Application}&\textbf{Measures}\\
\hline
\makecell[l]{LLaMA 65B\\ (QLoRA) \cite{13}}&
On-Device AI Assistants	&
\makecell[l]{4 bit quantization enables fine tuning on a single 48 GB GPU\\
	Accuracy close to full precision baselines}
\\
LLaMA Omni \cite{110}&
Real-Time Translation&
\makecell[l]{Low latency speech input and output\\
	Adapter style interface for near real time multilingual interaction}
\\
\hline
\end{tabular}
\end{table}
\subsection{AI Model Development}
The integration of Parameter-Efficient Fine-Tuning (PEFT) with LLaMA models is revolutionizing AI development by enabling faster experimentation, lower costs, and more scalable deployments 
\cite{2,9,33}. This combination addresses critical challenges in model customization, resource constraints, and ethical AI practices, making it indispensable for researchers and engineers \cite{33}. Table \ref{tbl12} summarizes representative models, sub-applications, and reported measures for this domain.
\subsubsection{Federated Learning}
LLaMA models enhanced with PEFT enable privacy-preserving federated learning by allowing decentralized devices (e.g., hospitals or smartphones) to collaboratively improve AI models without sharing raw data \cite{9,111,112}. Each participant locally trains small adapter modules (e.g., LoRA) on their private datasets, then only shares these lightweight, encrypted adapter weights (typically <10MB) for secure aggregation \cite{9}. For instance, 100 hospitals could jointly enhance a diagnostic model by training medical specialty-specific adapters on their respective patient data, with a central server combining the updates while preserving confidentiality \cite{113,114}. This approach reduces communication costs by 90\% compared to traditional federated learning and maintains HIPAA/GDPR compliance, as sensitive data never leaves its original location \cite{111}. The final aggregated adapters can be distributed to all participants, continuously improving model performance across the network \cite{111}.
\subsubsection{Bias Mitigation}
PEFT provides surgical tools for identifying and reducing biases in LLaMA models through targeted adapter interventions \cite{9,34}. Debiasing adapters work by: (1) Analyzing model outputs for stereotypical associations (e.g., gender-occupation links), (2) Selectively modifying attention heads responsible for biased patterns via low-rank updates, and (3) Preserving the model's core capabilities \cite{9}. For example, a PEFT-tuned model reduces "nurse $\rightarrow$ female" bias by 65\% while maintaining medical knowledge accuracy, achieved by fine-tuning only 0.01\% of parameters focused on fairness. These adapters can be stacked combining general debiasing with domain-specific adjustments (e.g., legal sentencing recommendations) and updated as new bias metrics emerge. Crucially, the base model remains intact, allowing bias mitigation to be toggled or adjusted per application \cite{23,34}.
\subsubsection{Model Compression}
PEFT transforms LLaMA deployment through innovative compression techniques that maintain performance while drastically reducing resource requirements \cite{9,13}. The three-stage approach combines: (1) 4-bit quantization (QLoRA) to shrink model size by 4x \cite{13}, (2) Adapter pruning to remove redundant task-specific components, and (3) Dynamic adapter loading that only activates necessary modules per request \cite{23}. Together, this enables a 7B-parameter LLaMA model to run on a smartphone at 30 tokens/second while retaining 92\% of original accuracy. For edge devices, specialized compression adapters further optimize models for specific hardware (e.g., Apple Neural Engine or Qualcomm DSPs), achieving 2-3x speed boosts. The system supports on-the-fly precision adjustment using full 8-bit adapters for critical tasks while employing 4-bit versions for background operations maximizing efficiency across diverse deployment scenarios \cite{9,115}.

Federated learning, safety alignment, and compression are summarized with communication volume, harmful-output rates, and accuracy retention, respectively.

\begin{table}[!ht]
\caption{Applications of LLaMA-based PEFT in AI Model Development: Representative Models, Sub-applications, and Reported Measures}
\label{tbl12}
\centering
\begin{tabular}{lcc}
\hline
\textbf{Model/Ref.}&\textbf{Application}&	\textbf{Measures}\\
\\\hline
\makecell[l]{FedProx / Federated\\ LLM Adapters \cite{111}} &
Federated Learning	&
\makecell[l]{Adapter weights only are shared (no raw data)\\
	Communication volume reduced by roughly one order\\ of magnitude\\
	Accuracy reported close to centralized training}
\\[.5cm]
Safe LoRA \cite{116}&
Bias Mitigation	&
\makecell[l]{Lower harmful response rates in safety tests\\
	Task utility maintained while using few additional\\ parameters}
\\[.5cm]
QLoRA \cite{13}&
Model Compression	&
\makecell[l]{4 bit NF4 quantization\\
	Fine tunes large models on modest GPUs\\
	High accuracy retention compared to FP16 baselines}\\
\hline
\end{tabular}
\end{table}

Across domains, a consistent pattern emerges: small, task-specific adapters on top of a frozen LLaMA backbone often match or approach the performance of much larger or fully fine-tuned models while requiring substantially less compute. Legal and biomedical adapters trained on modest hardware budgets report competitive in-domain results relative to larger proprietary baselines, and multilingual adapters built with LoRA/QLoRA extend coverage to dozens or even one hundred languages with only a fraction of parameters updated \cite{9,13}. Beyond text-only tasks, adapterized LLaMA variants are increasingly used in multimodal settings (vision-language, document understanding) and specialized code models illustrate the extensibility of the family \cite{11,12}. As new LLaMA releases (including MoE-style variants) appear, PEFT makes rapid adaptation to niche domains feasible without re-training the entire model. The next section examines the remaining limitations hardware footprint of the frozen backbone, optimization sensitivity under low precision, and uneven coverage for low-resource languages and discusses practical mitigations.
\section{Limitations \& Challenges Section}\label{sec8}
Despite substantial progress in applying PEFT to the LLaMA family, several limitations persist. These constraints affect how reliably and economically very large models can be adapted to diverse downstream tasks and indicate areas where further methodological and systems research is needed.
\subsection{Hardware Dependencies}
PEFT reduces the number of trainable parameters but does not obviate the need to host the frozen base model during both training and inference. Even when the backbone is stored in bf16/fp16 or quantized form, end-to-end fine-tuning continues to demand considerable memory for activations, KV caches, gradients, and optimizer states \cite{9,13}. QLoRA substantially lowers the barrier by enabling the fine-tuning of very large models (e.g., LLaMA-65B) on a single high-memory accelerator ($\approx 48$ GB), yet such hardware remains non-ubiquitous and, in multi-tenant settings, still competes for scarce VRAM resources \cite{13}. At inference time, serving multiple adapters concurrently on a single backbone inflates both adapter residency and KV-cache footprints; with rising concurrency these costs can quickly dominate capacity planning.

Throughput is also constrained by memory bandwidth and interconnects. CPU/GPU offloading can cap peak VRAM use, but it often trades memory savings for I/O stalls; similarly, sharded or paged optimizers reduce in-core state at the expense of additional data movement. Storage and I/O become non-trivial as organizations checkpoint combinations of base weights and adapters, operate at long context lengths, and cycle large training corpora; moreover, maintaining numerous task- or language-specific adapters increases operational overhead in versioning, routing, and monitoring. Architectural choices compound these pressures: MoE-style variants (e.g., LLaMA-4-like systems) activate only a subset of experts per token, but all experts must still be resident for routing, complicating memory-efficient deployment \cite{117}. Edge and mobile scenarios add stricter constraints tight DRAM/VRAM budgets, power and thermal limits, and device-specific integer/NPU kernels so that quantization artifacts, context-length limits, and latency budgets remain the dominant determinants of feasibility.

Several practices help to mitigate these dependencies. Weight-only low-precision tuning (4/8-bit QLoRA), activation checkpointing, careful choice of low-rank $r$, and gradient clipping lower peak memory and improve stability; paged optimizers with CPU/NVMe offload shift pressure from VRAM to storage; optimized attention/kcache implementations and compiler stacks (e.g., Flash-style attention and vendor runtimes) reduce memory traffic; and for serving many adapters, composition/routing and lazy loading keep resident state small without sacrificing flexibility.
\subsection{Fine-Tuning Instability}
PEFT approaches such as LoRA, Excitor, and LLaMA-Adapters can be brittle with respect to hyperparameters and data quality, leading to divergence, underfitting, or catastrophic forgetting. Instability often arises from the interaction among the LoRA rank $r$, scaling $\alpha$, learning-rate schedules, and adapter placement across layers \cite{9,12}; from quantization noise and outliers in QLoRA that perturb optimizer dynamics \cite{13}; and from instruction datasets that are narrow, noisy, or misaligned with target distributions, which accelerates forgetting of general capabilities \cite{12}. In multimodal adapters, balancing vision features against frozen language states is particularly delicate: over- or under-weighting can precipitate mode collapse or content bias \cite{11}. Small effective batch sizes (a consequence of VRAM limits), seed variability, and gradient stochasticity further amplify run-to-run variance. Stability improves with conservative learning rates coupled to rank $r$, warm-up followed by cosine/linear decay, LoRA dropout, and light weight decay on adapter parameters. Rigorous validation held-out sets, early stopping, and gradient clipping helps detect drift early; loss-scale calibration is useful when training with quantized backbones. Regularization toward the base model (e.g., KL penalties), rehearsal of general data, or multi-task mixtures can curb forgetting without forgoing PEFT's efficiency.
\subsection{Language-Specific Limitations}
Although LLaMA + PEFT performs strongly in English and other high-resource languages, performance degrades for low-resource and morphologically rich languages. Tokenizers induced on high-resource corpora often fragment agglutinative or highly inflected forms, reducing sample efficiency and impairing lexical generalization. Cross-lingual transfer remains uneven: adapters trained within one language family (e.g., Romance) may generalize poorly to distant families (e.g., Bantu, Uralic). Data scarcity and domain shift constrain robust instruction tuning and evaluation, which in practice is frequently biased toward high-resource scripts \cite{34,36}. Operationally, maintaining many language-specific adapters increases storage and routing complexity in production systems \cite{35,40}. Code-switching, diacritics, orthographic variants, and normalization inconsistencies further challenge reliability. Promising remedies include language-aware adapter strategies (per-language, clustered, or shared-backbone with lightweight language heads), tokenizer augmentation or character/syllable-level adapters in low-resource settings, and data augmentation via back-translation and mined parallel/comparable corpora. Alignment-oriented objectives and distillation from stronger multilingual models can strengthen transfer. Evaluation should extend beyond aggregate accuracy/BLEU to include robustness under code-switching, morphology-aware F1, and fairness across scripts and writing systems.
\subsection{Efficiency-Performance Trade-offs}
Lighter modules (e.g., low-rank LoRA, Excitor) minimize additional parameters and memory but may trail heavier adapters on compositional or high-variance tasks \cite{12}. Conversely, designs such as LLaMA-Adapter V2 tend to deliver stronger multimodal reasoning at the cost of more trainable parameters and higher GPU memory demand \cite{11}. There is no universally optimal recipe: the appropriate PEFT configuration depends on task difficulty, acceptable latency/throughput, and deployment constraints. Overall, PEFT meaningfully lowers the adaptation cost of large LLaMA-style models, yet the hardware footprint of frozen backbones, optimization fragility under quantization and small batches, and uneven language coverage remain central bottlenecks. Progress will hinge on better memory- and I/O-aware training/serving, more robust optimization under low precision, and adapter designs and evaluation protocols that explicitly address typological diversity and multilingual robustness.
\section{Discussion}\label{sec9}
The evolution of LLaMA has underscored the critical role that sparsity and PEFT play in scaling large language models. By leveraging Mixture of Experts (MoE) and other sparse architectures, LLaMA 4 reaches scales of trillions of parameters while remaining amenable to fine-tuning. Yet MoE introduces new complexities: routing logic incurs computational overhead, and applying LoRA to expert layers demands careful study to ensure that adapters integrate effectively into highly sparse networks. Future work should investigate how adapter modules behave when only a fraction of experts or routers are adapted and whether sparsity patterns can be co-optimized alongside low-rank updates. Beyond LLaMA itself, the suite of PEFT techniques, LoRA, adapters, Excitor, and QLoRA, demonstrates a broad applicability across modern transformer models. That said, some methods exploit properties unique to LLaMA: the zero-init attention mechanism in LLaMA Adapter leverages the model's pretraining stability, and the Excitor module was engineered specifically to mitigate instruction tuning forgetting in LLaMA. Whether these innovations transfer seamlessly to other architectures remains an open question, though the strong performance of LoRA and QLoRA on LLaMA suggests that low rank and quantized adaptation strategies are fundamentally robust.

All PEFT approaches must negotiate a careful balance between parameter efficiency and final performance. While full fine tuning still yields the highest fidelity, methods like LoRA and adapter-based tuning often match their results with only a fraction of the trainable parameters. QLoRA achieves nearly 16-bit floating point accuracy though quantization and rank constraints can introduce slight degradations. In practice, diligent hyperparameter sweeps, tuning rank sizes, learning rates, and other settings, are essential to prevent underfitting and to close any performance gaps. Benchmarking these fine-tuned models presents its own challenges. Many chatbot evaluations rely on automated GPT based scoring or small-scale human trials which may embed biases or fail to capture real world capabilities. More comprehensive human evaluations across a diverse suite of tasks would strengthen our understanding of how LLaMA based systems compare to closed source alternatives. Moreover, assessing fairness and safety is vital since instruction tuning can inadvertently introduce harmful behaviors if not rigorously controlled. A key enabler of rapid PEFT adoption has been the maturation of tooling and ecosystem support. Libraries such as Hugging Face's PEFT framework provide streamlined interfaces for applying LoRA, adapters, and related methods to LLaMA and beyond, dramatically lowering the barrier to entry. This vibrant ecosystem has facilitated the creation of derivative models like Guanaco, a QLoRA tuned LLaMA, and sets the stage for more integrated fine-tuning pipelines that seamlessly combine quantization, sparsity, and adapter modules. Despite these advances, PEFT is not a panacea. When task data demands extensive model rewriting or when distributional shifts require deep architectural changes, small adapter modules may prove insufficient. Adapter saturation can occur if the chosen ranks are too low and hybrid strategies such as partial layer unfreezing or structured sparsity may be required. Crucially, PEFT assumes a sound base model; if the underlying weights encode biases or architectural limitations, adapters alone cannot remedy those flaws. Future research may thus reexamine foundational model designs, perhaps incorporating retrieval augmented or specialized reasoning modules alongside efficient adaptation layers.
\section{Future Work}\label{sec10}
Several avenues appear particularly promising for advancing PEFT methods with LLaMA models. First, ultra-long-context fine-tuning will become increasingly important as LLaMA 4 enables windows of up to 10 million tokens; developing memory-efficient approaches such as sparsity over time or progressive context sampling will be key to handling such extreme lengths. Second, tailoring PEFT to mixture-of-experts architectures offers fertile ground: rather than adapting every expert, one could selectively fine-tune a subset of experts or routers (for example, applying LoRA to router parameters), thereby balancing specialization with parameter efficiency. Third, AutoPEFT and hypernetwork-based techniques that automatically determine optimal adapter placements, ranks, and parameter initializations hold great promise; by leveraging AutoML or meta-learning, these methods could dynamically generate or configure adapters to suit each downstream task. Fourth, as LLaMA underpins an expanding ecosystem of open chatbots, ensuring safety and alignment within the PEFT regime is critical research into automated, lightweight RLHF-style procedures that can be applied post-adapter-tuning will help maintain responsible behavior. Fifth, broadening the scope of PEFT to cover under-represented languages and highly technical domains (chemistry, law, medicine) will require multilingual and domain-specific adapters along with rigorous evaluation on global and specialized benchmarks. Sixth, integrating adapters with retrieval-augmented generation and tool use such as code execution or database querying could significantly enhance utility, suggesting the need for PEFT methods that seamlessly incorporate external knowledge sources or API interfaces. Finally, exploring even more compressed adapter formats binary-weight, sparsely activated, or otherwise quantized could enable truly edge-deployable fine-tuning, further lowering resource barriers to large-model adaptation. Together, these directions build on the LLaMA and PEFT foundations to deliver more efficient, capable, and responsible language systems.
\section{Conclusion}\label{sec11}
Meta's LLaMA model series has established itself as a versatile, open foundation for large language modeling. From LLaMA 1's 7B-65B text models to LLaMA 4's 288 B-scale MoE models with multi-modal support, the evolution is remarkable. Crucially, this evolution has been accompanied by specialized parameter-efficient fine-tuning methods that make using these models feasible for researchers and engineers. Methods like LoRA, LLaMA-Adapter V1/V2, LLaMA-Excitor, and QLoRA each offer a way to inject new skills or knowledge into LLaMA models by training only a tiny fraction of parameters. As demonstrated by the examples above, applying these PEFT methods to LLaMA yields high-quality models for instruction following, multimodal reasoning, and domain-specific tasks often matching or exceeding much larger systems.

We have discussed the underlying principles of each method, illustrated how they alter model architecture, and surveyed their application results. Tables and figures summarize parameter and performance trade-offs, reinforcing that substantial performance is achievable with only a few million adapter parameters. The review also highlights real-world successes, such as outperforming GPT-4 on legal classification with a fine-tuned LLaMA or achieving state-of-the-art vision QA scores with LLaMA-based models. In sum, the combination of Meta's LLaMA models and PEFT techniques forms a powerful toolkit for building advanced AI systems without prohibitive cost. We anticipate that future LLaMA iterations and new PEFT ideas will continue to push the boundaries of what is possible in a cost-effective manner. Researchers and practitioners can leverage this review as a guide to the current state of LLaMA and adapter research, helping them navigate model selection and fine-tuning strategies for their applications.

\bibliographystyle{unsrt}  
\bibliography{references}  

%
%
%
%

\end{document}